\documentclass{article}
\usepackage{iclr2027_conference,times}

\usepackage{amsmath,amsfonts,bm}

\def\eqref#1{equation~\ref{#1}}

\def\1{\bm{1}}

\DeclareMathAlphabet{\mathsfit}{\encodingdefault}{\sfdefault}{m}{sl}
\SetMathAlphabet{\mathsfit}{bold}{\encodingdefault}{\sfdefault}{bx}{n}

\usepackage{hyperref}
\usepackage{url}
\usepackage{booktabs}
\usepackage{graphicx}
\usepackage{amsmath}
\usepackage{amssymb}
\usepackage{microtype}
\usepackage{fvextra}
\usepackage{subcaption}

\usepackage[T1]{fontenc}
\usepackage{lmodern}
\usepackage{fvextra}

\DeclareUnicodeCharacter{2010}{-}     
\DeclareUnicodeCharacter{2011}{-}     
\DeclareUnicodeCharacter{2012}{-}     
\DeclareUnicodeCharacter{2013}{-}     
\DeclareUnicodeCharacter{2014}{-}     
\DeclareUnicodeCharacter{2212}{-}     
\DeclareUnicodeCharacter{03A3}{sum}   
\DeclareUnicodeCharacter{2265}{>=}    
\DeclareUnicodeCharacter{2264}{<=}    
\DeclareUnicodeCharacter{00B1}{+/-}   
\DeclareUnicodeCharacter{2192}{->}    
\DeclareUnicodeCharacter{00A0}{ }     
\DeclareUnicodeCharacter{FE0F}{}      
\DeclareUnicodeCharacter{20E3}{}      

\title{Hill Sampling for Test-Time Scaling:\\ A Simple and Better Alternative to\\ Repeated Sampling, Evolution, and Training}

\author{Jacob Beck, Philip V. Ogren, Ari Kobren \\
Oracle \\
\texttt{\{jake.beck,philip.ogren,ari.kobren\}@oracle.com}
}

\newcommand{\method}{Hill Sampling}

\newcommand{\modelname}[1]{\texttt{#1}}

\iclrfinalcopy
\begin{document}
\maketitle

\begin{abstract}
Large language models (LLMs) can improve solutions to verifiable scientific and algorithmic problems by spending additional computation at test time.
Recent systems achieve strong results with increasingly elaborate evolutionary search harnesses or by updating model parameters during test-time training.
We ask how much of this machinery is necessary.
We introduce \emph{Hill Sampling}, a form of hill-climbing optimization that repeatedly samples candidate programs from a frozen LLM, retains the best program found so far, and conditions all subsequent samples on that program.
We evaluate the method on circle packing, sums and differences of sets, and Erd\H{o}s' minimum-overlap problem using three open-weight models.
Hill Sampling sets a new state of the art on circle packing among published methods, improves over the AlphaEvolve reference on Erd\H{o}s' minimum-overlap problem, and achieves strong results on sums and differences of finite sets.
The circle-packing and Erd\H{o}s results require only hours of wall-clock time on eight NVIDIA H100 GPUs.
We compare Hill Sampling against what is, to our knowledge, the largest application by parameter count of evolution strategies (ES) to LLM weights at test time.
Surprisingly, when evaluating a method by the best program it generates, we find that learning model weights via ES is worse than invoking ES with a learning rate set to zero, i.e., using random weight-space perturbations to search for better models.
Moreover, repeated sampling outperforms both ES methods, and Hill Sampling is the strongest of all.
These results suggest a simple test-time compute allocation strategy: repeatedly sample edits to the best verified solution found so far before introducing additional complexity, such as adding archives, diversity mechanisms, evolutionary scaffolds, or test-time parameter learning.
\end{abstract}

\section{Introduction}
Practitioners using large language models (LLMs) are increasingly spending additional computation at test time to achieve better results.
For verifiable problems, an LLM can generate many candidate programs, evaluate them with an executable verifier, and use the feedback to improve subsequent candidates.
Systems such as FunSearch~\citep{romeraparedes2024funsearch}, AlphaEvolve~\citep{novikov2025alphaevolve}, and ShinkaEvolve~\citep{lange2026shinka} use complex harnesses based on this paradigm for mathematical and algorithmic \emph{discovery}, where the goal is to find the highest-scoring solution.

We ask how much of this complex machinery is necessary.
Starting from repeated sampling, we add only one persistent state: the best verified program found so far.
We call the resulting procedure \emph{Hill Sampling}, as it combines hill-climbing optimization and repeated sampling.
At every round, a frozen LLM generates new candidate programs conditioned on an incumbent.
If a candidate outperforms the incumbent, it becomes the new incumbent.
Unlike many existing systems, there is no archive, crossover, parameter update, or explicit diversity objective.
Our goal is not to establish the best sample efficiency in terms of evaluations or LLM completions; it is to understand whether a simple implementation can achieve state-of-the-art discovery with a practical wall-clock budget.

We evaluate Hill Sampling on three verifiable mathematical optimization problems using three open-weight models: circle packing with \modelname{gpt-oss-20b}, sums and differences of finite sets with \modelname{OLMo-3.1-32B-Instruct}, and Erd\H{o}s' minimum-overlap problem with \modelname{Mistral-Small-3.1-24B-Instruct}.
On circle packing, \method{} sets a new state-of-the-art result among published methods in under five hours on eight NVIDIA H100 GPUs.
On Erd\H{o}s, it improves on the AlphaEvolve algorithmic-discovery reference in 12 hours.

The simplicity of Hill Sampling also lets us ask a complementary question raised by recent work on the geometry of pretrained weight spaces.
\citet{gan2026thickets} argue that useful task-specific experts can be found densely around the weights of sufficiently large pretrained models, motivating random perturbation of model weights and selection or ensembling.
We therefore test whether similar exploration via model noise can help verifiable program discovery, where candidate solutions can be evaluated exactly.
We implement what is, to our knowledge, the largest-scale study by parameter count of evolution strategies (ES) applied directly to LLM weights in this setting.
Additionally, we implement a model-noise variant by setting the learning rate of ES to zero.

The resulting progression is informative.
ES learning often improves mean return while reducing the maximum return.
Setting the ES learning rate to zero removes the learning step but still searches via random weight perturbations.
Those perturbations can improve discovery, but repeatedly sampling from the unperturbed model with ordinary token sampling is stronger still. 
This repeated sampling, in turn, is substantially improved by conditioning on the best program found so far. That is, \method{} outperforms all other methods.

Our contributions are:
\begin{itemize}
    \item We introduce and evaluate \method{}, a minimal best-so-far conditioned sampling procedure for test-time program discovery, and show that it reaches state-of-the-art algorithmic-discovery performance on one problem and strong results on the other two with practical wall-clock cost.
    \item We establish the largest ES training pipeline to date, and demonstrate that setting the learning rate to zero can actually improve maximum return over the course of training, even when using mechanisms to prevent entropy collapse.
    \item We show that token-level sampling is a stronger source of useful diversity than random model perturbations in our verifiable discovery setting. 
    \item We evaluate a broad range of mechanisms to encourage diversity, enable combinations of solutions, add exploration, and provide code execution results, and show them all to be unnecessary.
\end{itemize}

\section{Related Work}
\paragraph{LLM-guided program evolution.}
FunSearch established a general recipe for pairing LLM-generated program mutations with executable evaluation and retaining high-scoring programs~\citep{romeraparedes2024funsearch}.
While an early approach, FunSearch already included complex components, such as sub-populations in islands to maintain diversity, sampling candidates relative to evaluation performance, and multiple prior solutions given in context for the purpose of recombination.
AlphaEvolve scales this pattern and reports strong results on mathematical and systems problems~\citep{novikov2025alphaevolve}.
ShinkaEvolve focuses on sample efficiency, using complex parent selection, prompting with multiple parents, sub-populations, novelty rejection, and other mechanisms to reduce the number of samples needed for good discoveries~\citep{lange2026shinka}.
Recent work has continued to develop sophisticated frameworks for LLM-guided evolutionary and scientific discovery~\citep{zheng2026dreamrsi,assumpcao2025codeevolve,yan2026pacevolve,jiang2026deltaevolve,liu2026evox,ye2026structured,wang2025thetaevolve}.
Our objective is different: we deliberately favor the simplest implementation that reaches strong or state-of-the-art discovery performance at a reasonable wall-clock cost rather than optimizing sample efficiency.

\paragraph{Simple discovery baselines.}
The concurrent work of \citet{gideoni2026simple} compares code evolution with repeated sampling, where the model is asked to solve the program from scratch many times, and sequential conditioned sampling, where each generation conditions randomly on prior solutions that executed successfully, with optional resets and phased evaluation.
They find simple methods to be competitive, but their simple baselines do not consistently surpass both AlphaEvolve and ShinkaEvolve on any of the domains they evaluated, and are worse on the three domains we evaluate.
They also report that domain knowledge can materially affect search, including on circle packing.
Our experiments qualify this picture: on circle packing, removing the initial domain-specific prompt and code does not affect our results, whereas on Erd\H{o}s it does reduce performance.


\citet{gupta2026harness} study many discovery harnesses and conclude that there is no universally best fixed harness.
They find useful behavior from some components, while other mechanisms are not consistently beneficial.
Therefore, they advocate for a complex adaptive allocation across harnesses.
Additionally, they do not ever match or exceed the scores from AlphaEvolve.
In contrast, our proposed method is simpler and does achieve competitive performance.

\paragraph{Test-time training and objective mismatch.}
EvoTune continuously refines the LLM with reinforcement-learning updates from solutions produced by evolutionary search~\citep{surina2025algorithm}, while TTT-Discover adapts the language model during inference using experience generated from search~\citep{yuksekgonul2026tttdiscover}.
These approaches make test-time training part of the discovery loop, whereas Hill Sampling keeps the model fixed and changes only the program state.
We compare to the addition of learning, even in conjunction with Hill Sampling, and find it to be an unnecessary complication.

For repeated generation, conventional reinforcement learning optimizes expected reward, which in the binary case corresponds to pass@1 and can favor safe, homogeneous outputs over the diversity needed for strong pass@$k$.
Pass@K Policy Optimization (PKPO) instead directly optimizes a joint objective over $k$ responses and derives unbiased, lower-variance estimators by drawing a larger batch of $n\geq k$ responses and averaging over its size-$k$ subsets~\citep{walder2025pkpo}.
Our ES (max@8) baseline targets the same maximum-over-$k$ objective, but deliberately uses the simpler maximum of eight responses per perturbed model rather than PKPO's larger-batch combinatorial estimator.
The latter would spend more of our fixed generation budget on each perturbed model and therefore reduce the number of independent weight perturbations we can evaluate.
Thus max@8 tests whether aligning the ES objective with discovery is beneficial, without importing the additional estimator complexity of PKPO.

\paragraph{Weight-space search and Neural Thickets.}
Evolution strategies (ES) provide a gradient-free, highly parallelizable way to optimize neural-network parameters from scalar rewards~\citep{salimans2017es, qiu2026evolution}.
We instantiate the largest full-parameter adaptation using ES to our knowledge to date, adapting 20-billion-parameter to 30-billion-parameter models.
While we are able to train these models to improve mean return, with a steady learning curve, the max performance does not improve, and is worse than setting the learning rate to zero, which simply adds model noise.

Relatedly, Neural Thickets argues that, for sufficiently large pretrained models, diverse task-specialized experts can occupy a dense neighborhood around the pretrained weights, motivating random weight-space search and selection~\citep{gan2026thickets}.
That work evaluates this phenomenon through downstream-task performance and ensembling, where generalization to a target task is central.
We ask a different question: whether weight-space perturbations are useful when the model responses can be evaluated exactly, with no generalization needed.
Our results show that, in these settings, token-level sampling in program space is more effective than either learned or fixed random weight perturbations.
Moreover, both of these are outperformed by Hill Sampling.

\section{Methods}
\subsection{Problem setting}
We consider optimization problems with an executable verifier.
A state $x$ is a program, and executing it produces a scalar reward $r$, which may be stochastic because the program itself can contain randomness.
An LLM defines a distribution over edited solutions $y\sim p_\theta(\cdot\mid x;T)$ at decoding temperature $T$.
The objective of test-time discovery is the maximum observed verified reward found within a fixed number of LLM edits ($M$),
\begin{equation}
    r^* = \max_{i\leq M} r_i.
\end{equation}
This objective differs from improving the average quality of samples: an update can increase expected reward while reducing the probability of a rare, exceptionally good discovery.

\subsection{Hill Sampling}
Let $x_0$ be the initial program, $x_t$ be the current program, called the \textit{incumbent}, and $r^*_t$ be the best reward observed so far at round $t$.
Each round, Hill Sampling draws $N$ edits independently from the current incumbent,
\begin{equation}
    y_{t,1},\ldots,y_{t,N}\sim p_\theta(\cdot\mid x_t;T),
\end{equation}
and executes each candidate once to obtain rewards $r_{t,1},\ldots,r_{t,N}$.
The incumbent $x_t$ is not re-evaluated; its previously observed reward $r^*_t$ is retained.
Let $i^*=\arg\max_i r_{t,i}$ index the best edit this round.
We then set
\begin{equation}
    (x_{t+1},r^*_{t+1})=
    \begin{cases}
        (y_{t,i^*},r_{t,i^*}) & \text{if } r_{t,i^*} \ge r^*_t,\\
        (x_t,r^*_t) & \text{otherwise}.
    \end{cases}
\end{equation}
Thus $r^*_{t+1}\geq r^*_t$ by construction, even when executing the same program can produce different rewards, and every accepted improvement immediately becomes the context for subsequent responses.
We accept new incumbents with equal reward to promote potential diversity.
There is no archive, diversity objective, crossover, or parameter update; the only persistent search state is the incumbent program $x_t$ and its stored best observed reward $r^*_t$.
This makes Hill Sampling a simple evolutionary algorithm whose knobs are simply the temperature, the number of rounds, and the number of samples drawn each round.

\paragraph{Hill Sampling (HS) 64 and 512.}
Hill Sampling ($N=64$) evaluates 64 edits per round, while Hill Sampling ($N=512$) evaluates 512.
Both use the same total sampling budget: Hill Sampling (64) runs for the specified number of rounds, while Hill Sampling (512) runs for eight times fewer rounds, trading more frequent incumbent updates for greater parallelizability.
For both, we instantiate eight vLLM instances, one per H100 GPU.
For $N=64$, each instance generates one response at a time until 64 responses are collected; for $N=512$, we generate 64 batches of eight responses.
To ensure sampling is invariant to batching, we set the respective vLLM environment variable (\texttt{VLLM\_BATCH\_INVARIANT}) and manually assign a unique seed to every response, incrementing seeds globally across batches and responses in the batch.

\section{Experimental Setup}
\subsection{Domains}
We study three verifiable mathematical optimization domains used in recent LLM discovery work \cite{novikov2025alphaevolve,lange2026shinka,yuksekgonul2026tttdiscover}, using the open-source implementation from \citet{openevolve}.
We measure results over three seeds per method, on each domain, and tune over multiple temperatures.
We arbitrarily divide our three models across the three domains, rather than evaluating each model on each domain, to enable reasonable computational constraints.
See Appendix~\ref{app:problems} for formal definitions, and Appendix~\ref{app:hyperparams} for hyperparameter details.
We additionally strengthen the validation functions by adding value and type checks to prevent observed reward hacking, with details in Appendix~\ref{app:evaluator}.
All compute is matched by the number of LLM completions.

Note that runs are sensitive to both the total number of rounds and execution timeout.
We fixed the total number of rounds (and therefore LLM completions) in early experiments in order to give the mean return of the Evolution Strategy method time to begin to plateau, which also resulted in significant diversity in the total runtimes between domains.
We find that increasing the code-execution timeout also can improve performance, with a particularly large improvement when increasing it from 5 to 20 seconds.
Even at a 20-second timeout, wall-clock time remains primarily bottlenecked by LLM generation rather than evaluation.
However, because of the volume of experiments, we use a 5-second execution timeout by default.
Code that passes this first evaluation is then re-evaluated with a 10-second timeout.
We later find that evaluating each program only once does not significantly affect performance.

\textbf{Circle packing} (Circles) asks for 26 non-overlapping circles contained in the unit square, with the objective of maximizing the sum of their radii.
For the standard circle-packing setting, the initial prompt and code come from phase two of a phased prompt schedule \cite{openevolve}.
Our \textbf{no initial information (NI)} variant removes this initial prompt and code, providing only the function signature and evaluation code as context.
The circles task uses \modelname{openai/gpt-oss-20b}, for 200 rounds (for N=64).

\textbf{Sums and differences of finite sets} (Sets) concerns the largest exponent $C_6$ governing how large a difference set $A-B$ can be relative to a controlled sumset $A+B$.
Following the computational formulation used in prior discovery work, programs construct a finite set $U$ of non-negative integers to maximize a given quantity.
The sets task uses \modelname{OLMo-3.1-32B-Instruct}, for 600 rounds (for N=64).

\textbf{Erd\H{o}s' minimum-overlap problem} (Erdos) asks for the smallest achievable worst-case overlap between a function and its complement, equivalently yielding an upper bound on the constant $C_5$.
The \textbf{no initial information (NI)} variant removes the initial solution code and retains only the function signature and evaluation code as context.
The Erdos task uses \modelname{Mistral-Small-3.1-24B-Instruct-2503}, for 80 rounds (for N=64).

\subsection{Baselines}
We compare Hill Sampling against other methods that use test-time compute.
\textbf{Repeated Sampling (RS)} repeatedly draws candidate programs from a frozen model as edits to the original code, without carrying the best program forward as the next editing state.
\textbf{Model Noise (MN)} evaluates 64 perturbed models, $\theta_i=\theta+\sigma\epsilon_i$, where we independently sample $\epsilon_i\sim\mathcal{N}(0,I)$ for $i=1,\ldots,32$ and set $\epsilon_{i+32}=-\epsilon_i$, as in antithetic sampling.
MN (64) evaluates one generation per model, while MN (512) evaluates 8 responses per model.
\textbf{Evolution Strategies (ES)} evaluates the same antithetic population and updates the underlying model using the scalable ES estimator of \citet{salimans2017es}, $\theta\leftarrow\theta+\frac{\alpha}{32\sigma}\sum_{i=1}^{32}\tilde{R}_i\epsilon_i$, where $\alpha$ is the learning rate and $\tilde{R}$ denotes standardized antithetic reward.
As in \citet{qiu2026evolution}, we store only the random seed and reward, reproducing the noise vector each time from the random seed, to save space, and undo each perturbation by subtracting it, rather than re-loading the model from disk. (While this does cause rounding errors, we find it to be faster and perform similarly.)
\textbf{ES (max@8)} evaluates eight responses from each perturbed model and assigns that perturbation their maximum reward, targeting the same set-level maximum objective as in \citet{walder2025pkpo}. 
\textbf{ES (softmax)} replaces standardized scalar weighting with a softmax over population rewards, increasing emphasis on the best perturbations, as in \citet{yuksekgonul2026tttdiscover}.
\textbf{\method{} + ES} uses \method{} in conjunction with ES (max@8) updates on the model.

For the Sets task, we use only the baselines that are batched, due to computational limitations: \textbf{RS, ES (max@8), \method{} (512) + ES}.
For the broader Erdos evaluation, we compare additional methods described in that section.
Additional details are in Appendix \ref{app:methods}.

\begin{figure}[tb]
\centering
\begin{subfigure}[t]{0.32\linewidth}
    \centering
    \includegraphics[width=\linewidth]{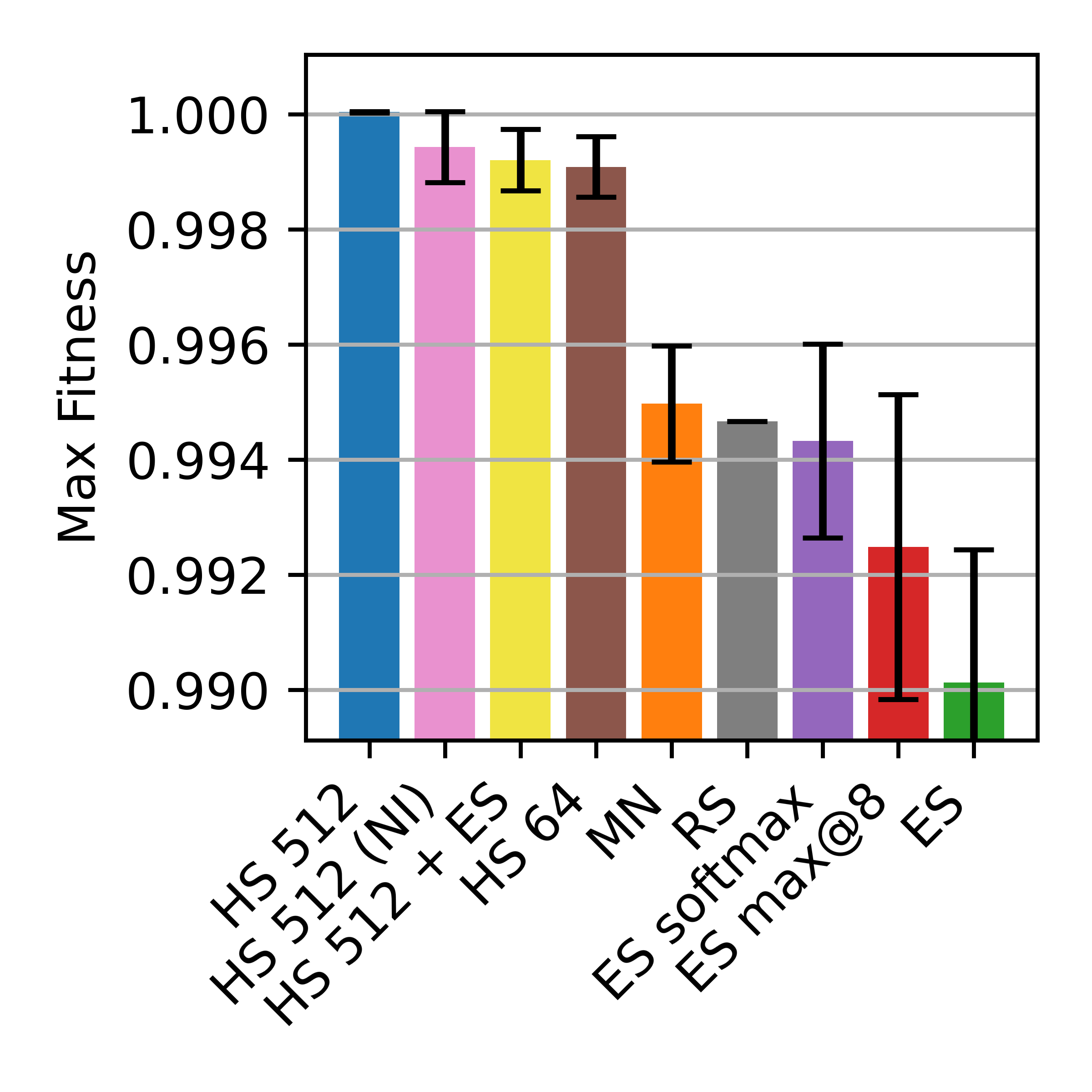}
    \caption{Circle packing.}
    \label{fig:main_circles}
\end{subfigure}
\hfill
\begin{subfigure}[t]{0.32\linewidth}
    \centering
    \includegraphics[width=\linewidth]{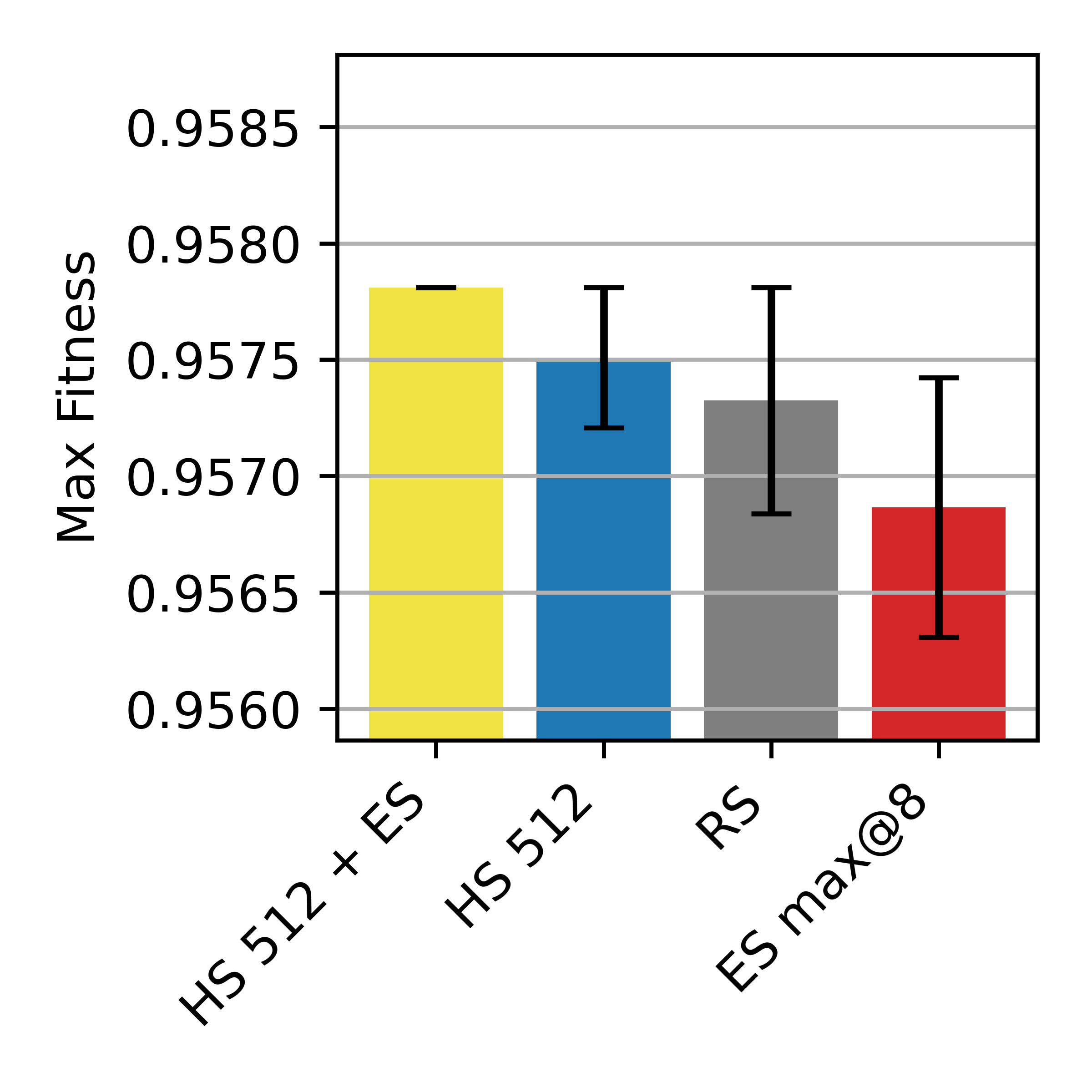}
    \caption{Sums and differences of sets.}
    \label{fig:main_sets}
\end{subfigure}
\hfill
\begin{subfigure}[t]{0.32\linewidth}
    \centering
    \includegraphics[width=\linewidth]{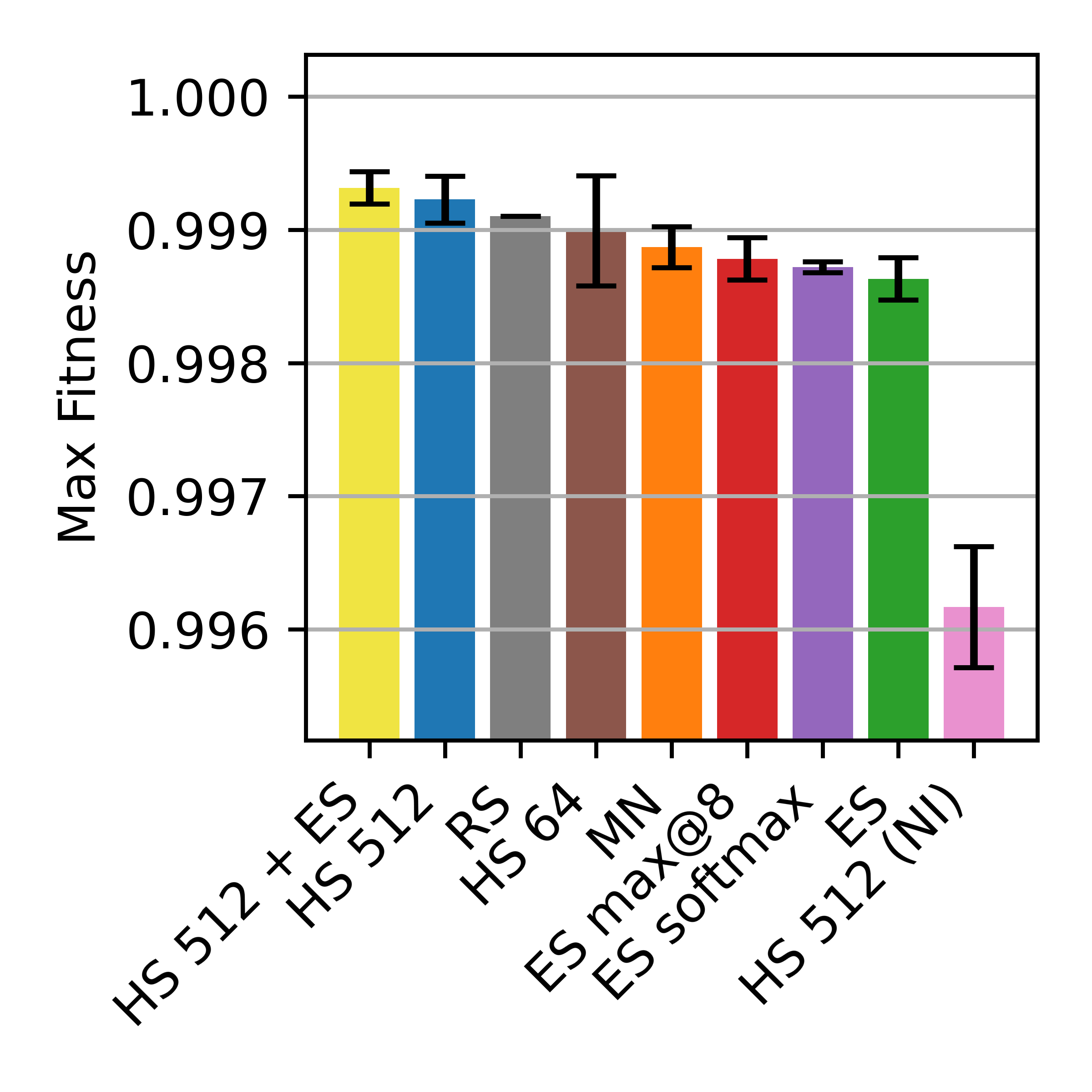}
    \caption{Erd\H{o}s minimum overlap.}
    \label{fig:main_erdos}
\end{subfigure}
\caption{\textbf{Main results.} Results are shown using scores normalized to AlphaEvolve.
Error bars, as in all subsequent plots, show standard error. 
On Circles, only variants of Hill Sampling achieve top performance, with ES performing the worst. 
Both Hill Sampling (512) and its NI variant with no initial information have a seed that achieves the top score. 
On Sets, there is no significant difference between the methods, with all methods having a seed that achieves the same top score, and ES performing the worst on average.
On Erdos, Hill Sampling (512) and Hill Sampling (512) + ES perform best.
RS achieves a high average max return, with no single seed performing as well as Hill Sampling (512).
Here, all ES variants underperform, and our NI variant is the worst, indicating a need for domain knowledge.}
\label{fig:main}
\end{figure}

\section{Results}

\begin{table}[tb]
\caption{\textbf{Best-result summary.} 
We report the best scores achieved by Hill Sampling (512) compared to existing work. The best result is in \textbf{bold}, and the second best is \underline{underlined}. 
Hill Sampling sets a new state of the art on the Circles task, among published methods, in under five hours, and beats AlphaEvolve on Erdos. 
Time and round of discovery are reported.
The Circles result used a single 100s timeout, Sets used two 5s timeouts, and Erdos used two 20s timeouts.
The number of LLM completions can be computed as 512 times the number of rounds.
Normalized scores are reported as a fraction of AlphaEvolve's score, or the reciprocal, such that higher is better.
}
\label{tab:main}
\centering
\small
\resizebox{\linewidth}{!}{%
\begin{tabular}{lccccccc}
\toprule
Domain & Our best & Our best normalized $\uparrow$ & AlphaEvolve & Best prior published & Time & Rounds & Model \\
\midrule
Circles $\uparrow$ & $\mathbf{2.635983084917604}$ & $1.0000456505194$ & $2.6358627564136983$ & $\underline{2.6359830774}$ (ThetaEvolve) & 4:33 & 14 & gpt-oss-20b \\
Sets $\uparrow$ & 1.109543 & 0.9578097426685801 & $\underline{1.158417281556896}$ & $\mathbf{1.21}$ (Hyra) & 1:13 & 2 & OLMo-3.1-32B \\
Erdos $\downarrow$ & $\underline{0.38089767}$ & $1.0000665904$ & $0.38092303510845016$ & $\mathbf{0.380876}$ (TTT-Discover) & 12:00 & 70 & Mistral-24B \\
\bottomrule
\end{tabular}%
}
\end{table}

Figure~\ref{fig:main} summarizes the primary comparisons.
\method{} is the strongest method on circle packing, including without the initial domain-specific information, where it reaches the same best score as the informed setting.
This is surprising, given that the default Circles problem presents the most domain specific information.
On Erdos, \method{} is also the strongest method, while adding ES does not improve it and removing the initial information substantially hurts performance.
On the Set problem, all methods perform similarly.

Table~\ref{tab:main} summarizes the best results achieved.
In under five hours, \method{} sets a new state of the art on circle packing, among published methods, evaluated without slack\footnotemark.
\method{} improves over the AlphaEvolve result on Erdos, while remaining under \citet{yuksekgonul2026tttdiscover}, who use the much larger \modelname{gpt-oss-120b}.
Note that even switching from \modelname{Mistral-24B} to \modelname{gpt-oss-20b} improved our result to 0.38088232 (1.0001068894978207 normalized).
Our Sets result remains below AlphaEvolve, new autonomous discoveries \citep{lin2026settling}, and constructions discovered (but not necessarily instantiated in memory) with human assistance \citet{lin2026settling, gerbicz2025sums,zheng2025sums}.

\footnotetext{AlphaEvolve reports scores without any slack (i.e., overlapping space) allowed between circles. 
We follow this convention for a fair comparison.
Some existing work \citep{lange2026shinka} evolves with slack in overlap and then removes it.
We followed that procedure for all other experiments in the paper.
However, doing so generally resulted in a lower score of 2.6359830848923913 after removing $10^{-12}$ from the radii, as the lowest integer power of 10 that still passes the verifier.
We report the result from ThetaEvolve \citep{wang2025thetaevolve} in the table, as the best zero-slack result published in a prior paper.
Two concurrent web results have surfaced with undocumented methods during writing that claim better results: Hyra \citep{lin2026hyra_n26} and Evölther \citep{gotherlabs2026circlepacking}.
We find that both do not pass our verifier. 
Removing slack in the same way, we find scores of 2.6359830691068447 (worse) and 2.6359830849176054 (better), respectively.
While Evölther achieves the greatest value we can verify, at this level of precision, validity can depend on floating-point evaluation details, including the order of operations.
Our results should be read with these qualifications in mind.
%
}

While \method{} achieves about 95\% of the AlphaEvolve score on Sets, so do the other baselines evaluated, which is suggestive of a model or domain knowledge failure.
Inspecting the initial code from AlphaEvolve, the initial code is misleading: 
the given solution searches within the space of integers below 250, while constructions in the original paper that proposed the problem, and the best known solution,
all produce sets that are sparse, with a maximum integer potentially in the millions.
In a separate run, using \modelname{gpt-oss-20b}, we run 150 rounds with no initial code, use a single 20-second timeout, and include a 
note in the prompt that the known solution is sparse, with tens of thousands of elements over a range of hundreds of thousands to millions\footnote{We do not include this result in Table~\ref{tab:main}, given the increased domain knowledge in the task.}.
That run achieved 99.02\% of the AlphaEvolve score, suggesting improvement from the model and/or domain knowledge.

\subsection{Model noise: weight-space exploration is not enough}
\begin{figure*}[tb]
\centering
\begin{minipage}{0.49\textwidth}
  \centering
  \includegraphics[width=\linewidth]{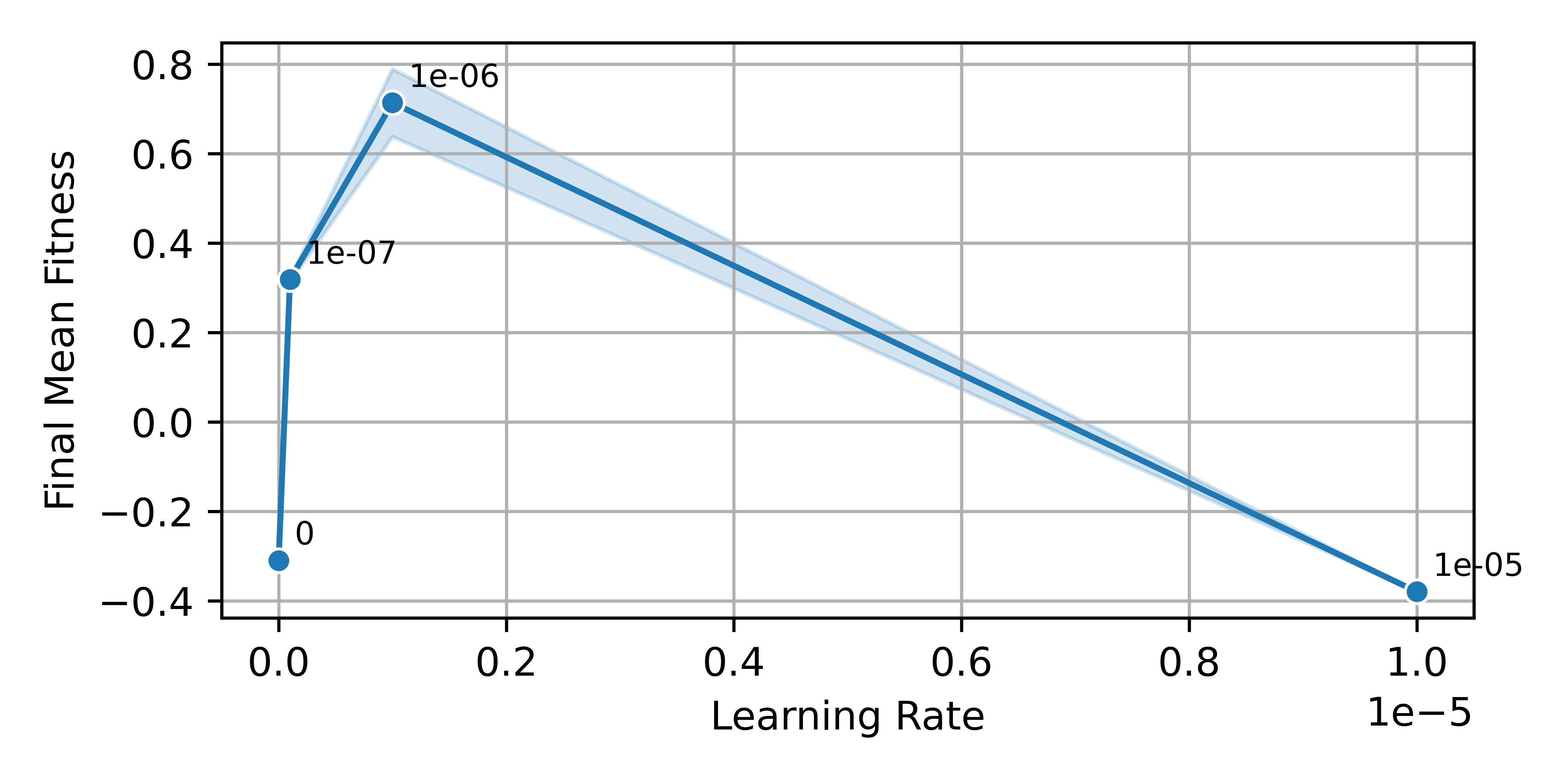}
  \small (a) Final mean fitness versus ES learning rate.
\end{minipage}
\hfill
\begin{minipage}{0.49\textwidth}
  \centering
  \includegraphics[width=\linewidth]{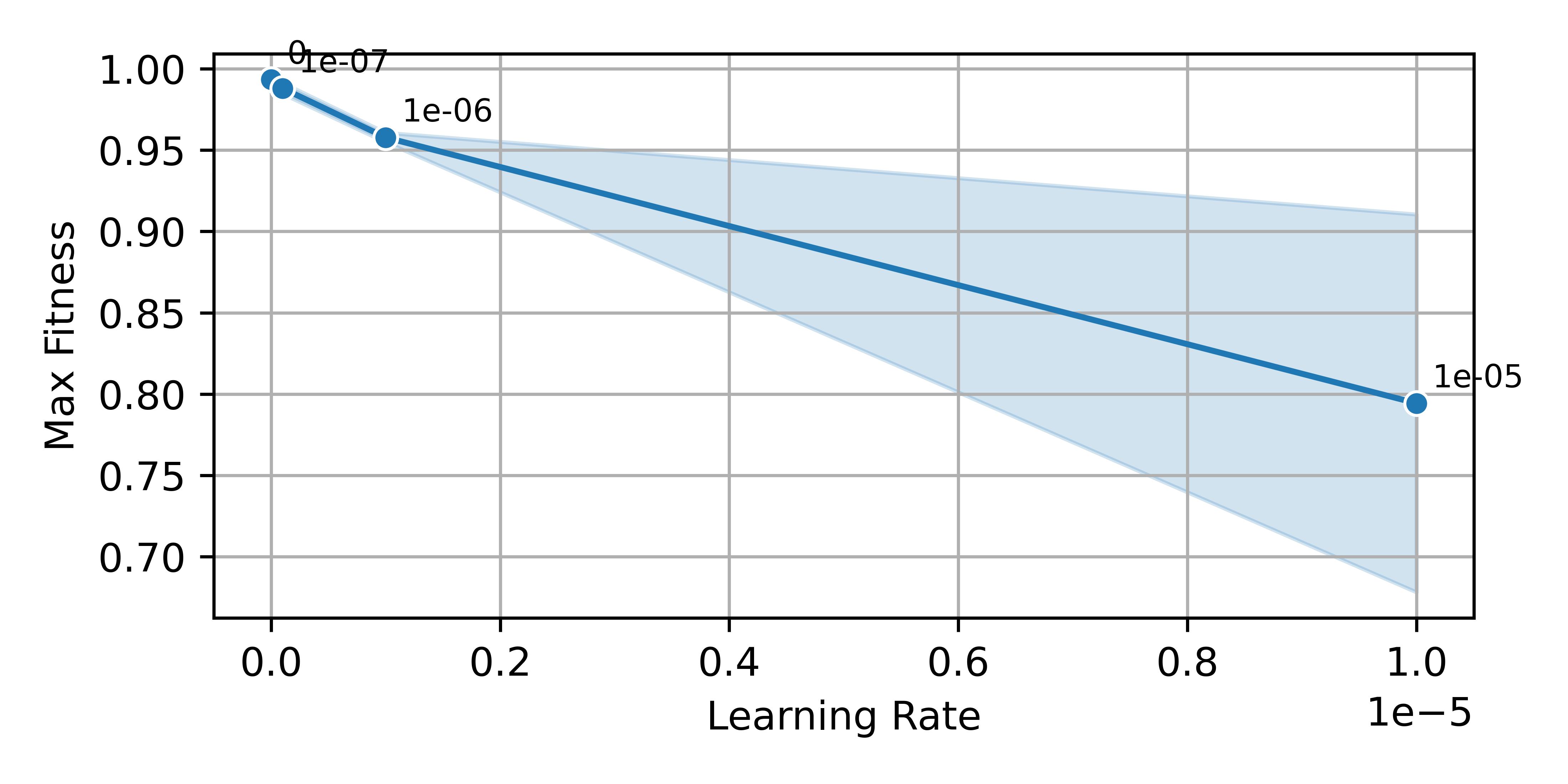}
  \small (b) Maximum fitness versus ES learning rate.
\end{minipage}
\caption{\textbf{ES improves mean return but not maximum return.} With $\sigma=10^{-3}$ and temperature 0, a nonzero ES learning rate improves final mean fitness (averaged over the last 10\% of data) while the best maximum fitness occurs at learning rate zero, on the Circles task. Thus the ES update can improve average quality while degrading the discovery objective.}
\label{fig:es_lr}
\end{figure*}

Our first weight-space experiment asks whether ES learning improves the discovery objective.
Figure~\ref{fig:es_lr} shows the key mismatch on circle packing: increasing the ES learning rate improves the final mean return, but the maximum return is best at learning rate zero.
Standard ES improves the average quality while harming the extreme statistic that matters for discovery.
This motivates setting the learning rate to zero, leaving only random model perturbations.

\begin{figure*}[tbh]
\centering
\begin{minipage}{0.49\textwidth}
  \centering
  \includegraphics[width=\linewidth]{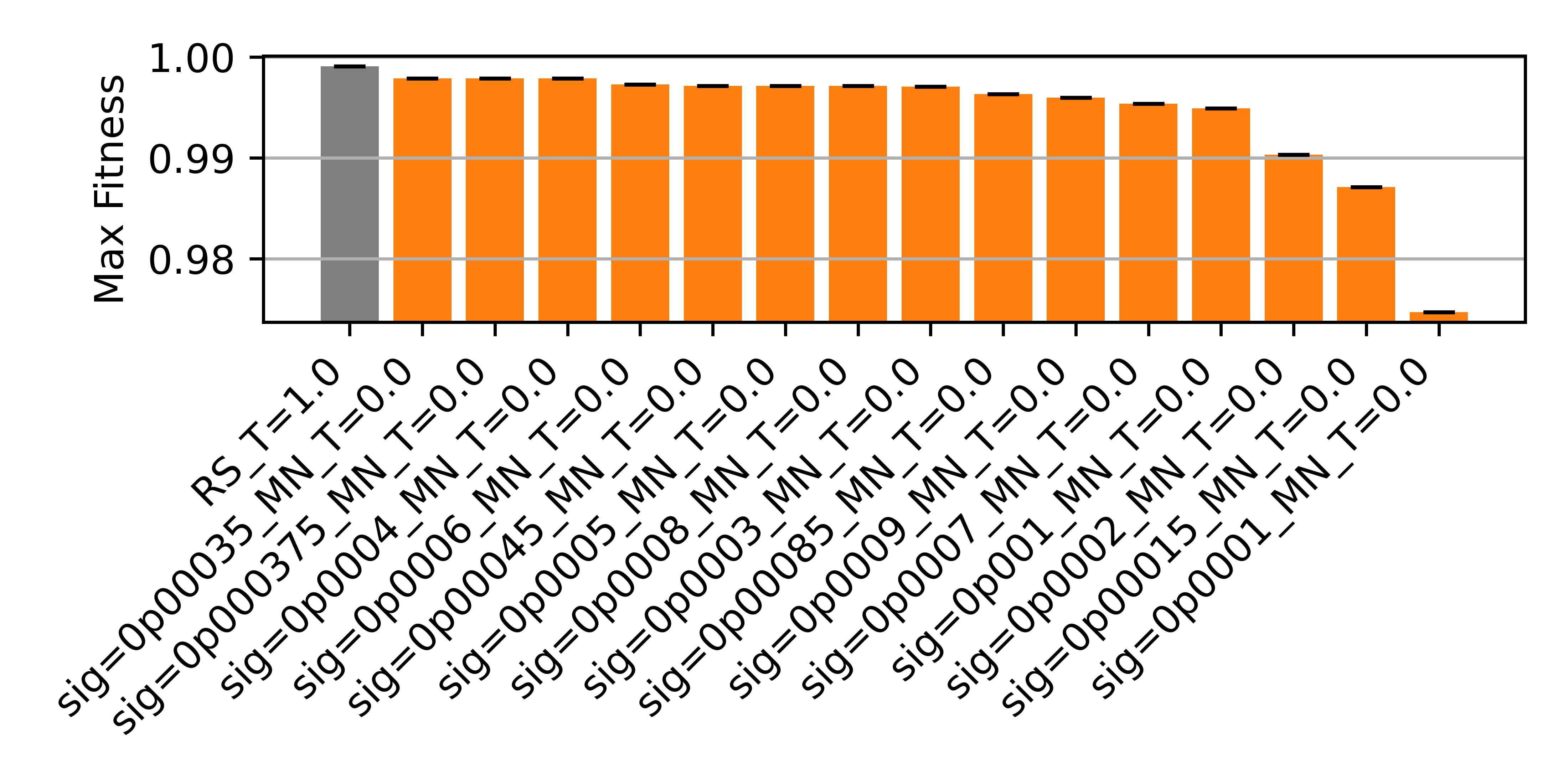}
  \small (a) Erd\H{o}s: maximum fitness.
\end{minipage}
\hfill
\begin{minipage}{0.49\textwidth}
  \centering
  \includegraphics[width=\linewidth]{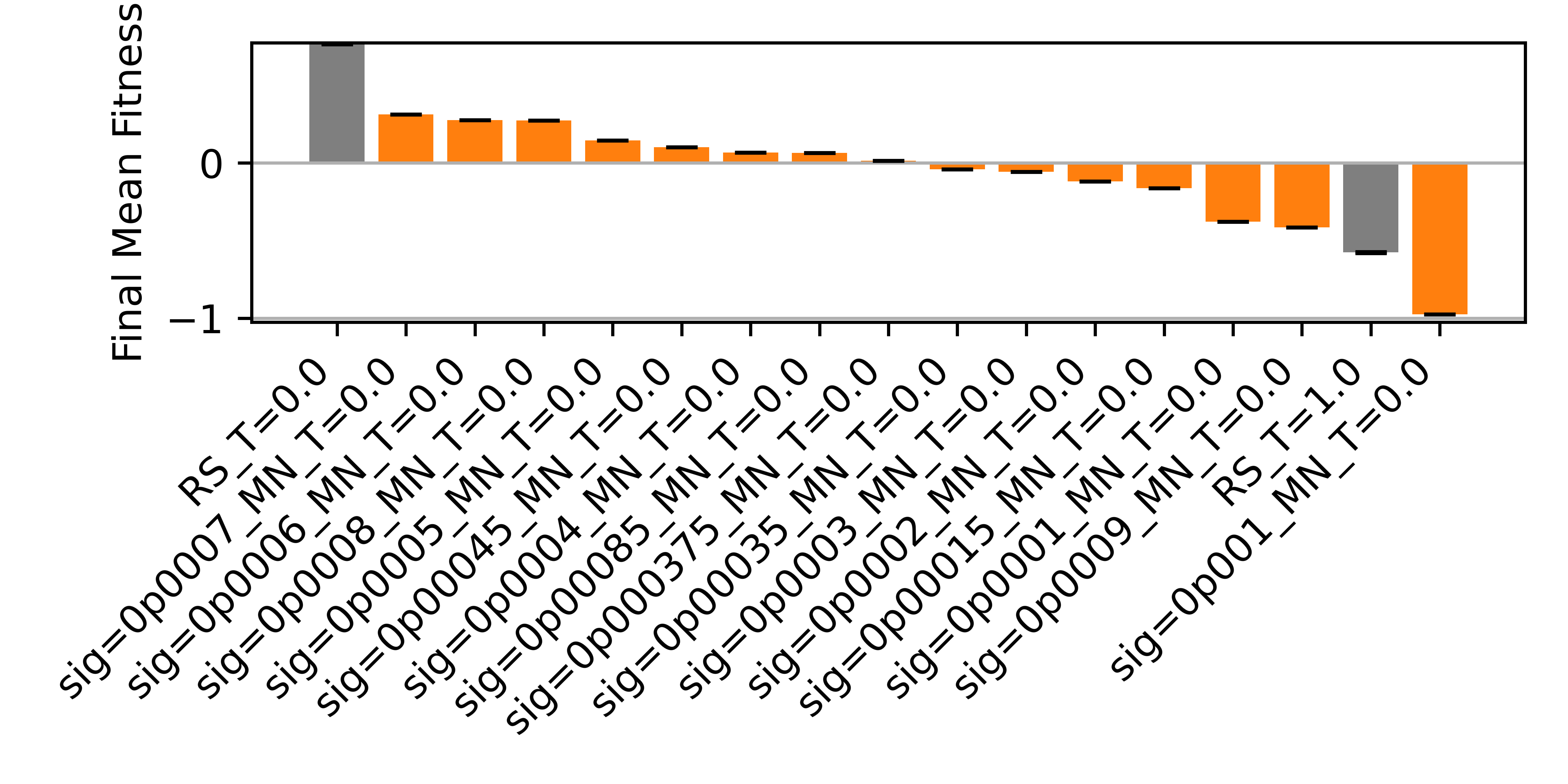}
  \small (b) Erd\H{o}s: final-10\% mean fitness.
\end{minipage}
\vspace{0.5em}
\begin{minipage}{0.48\textwidth}
  \centering
  \includegraphics[width=\linewidth]{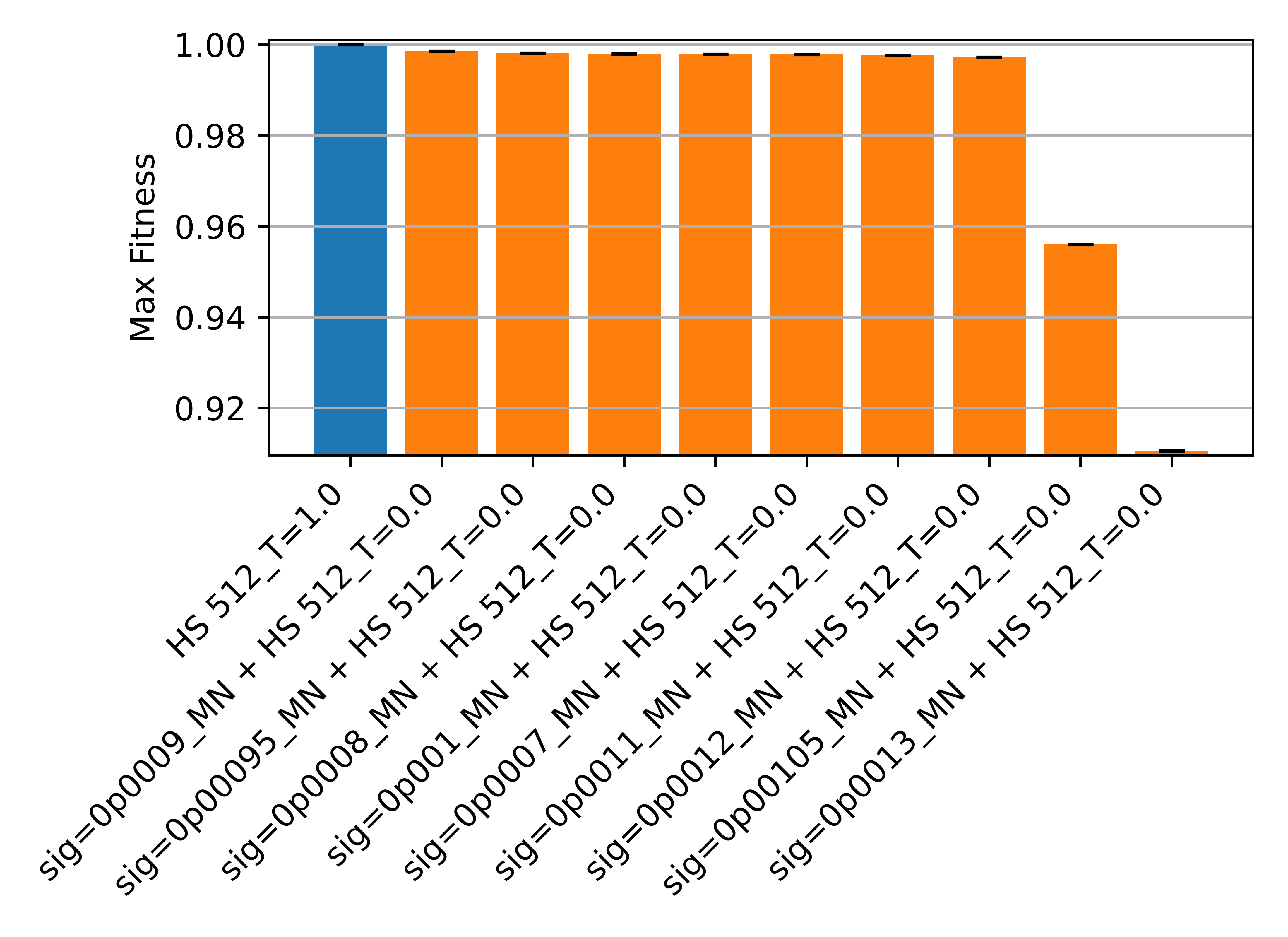}
  \small (c) Circles: maximum fitness.
\end{minipage}
\hfill
\begin{minipage}{0.48\textwidth}
  \centering
  \includegraphics[width=\linewidth]{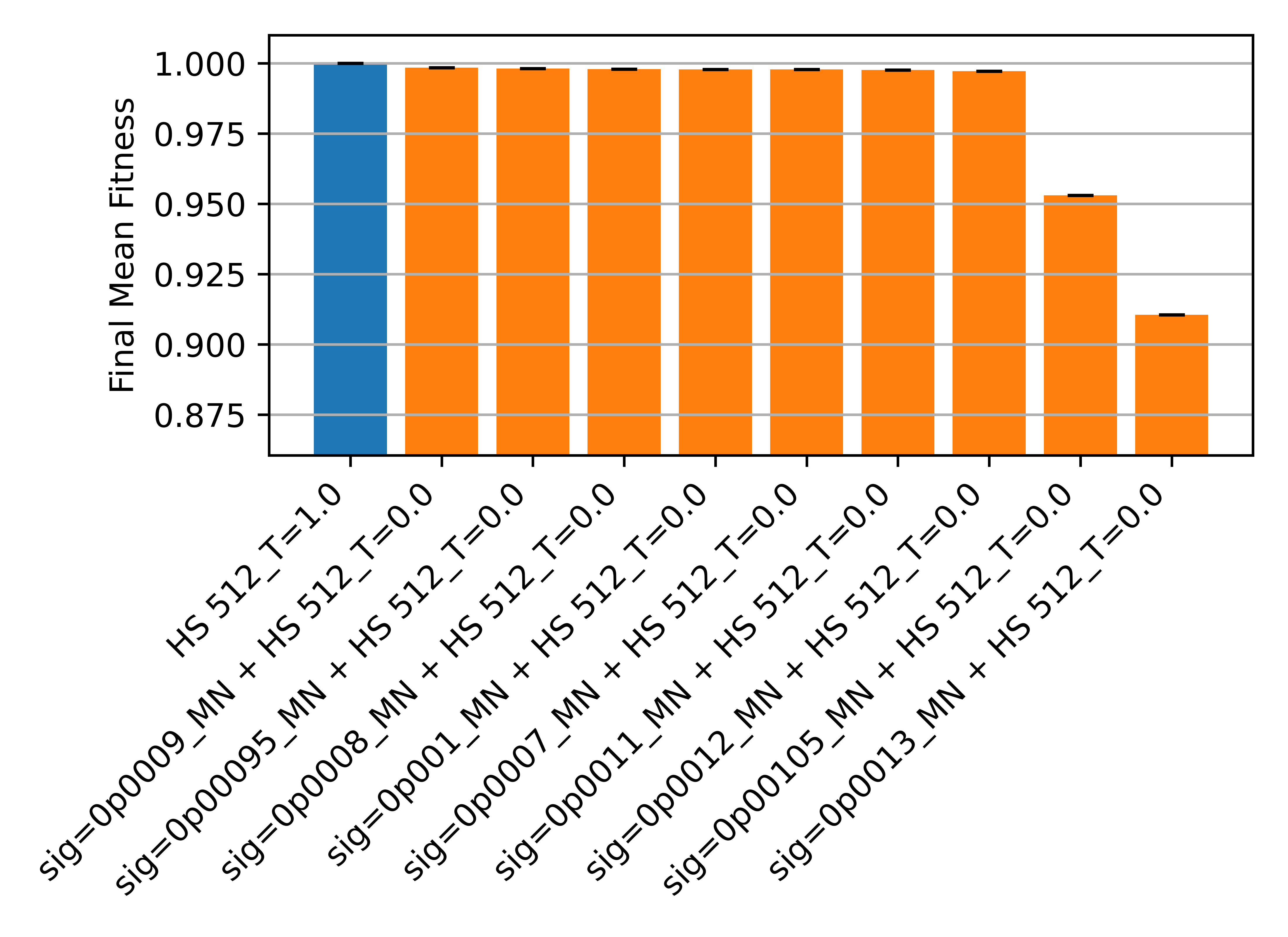}
  \small (d) Circles: final-10\% mean fitness.
\end{minipage}
\caption{\textbf{Model noise is weaker than ordinary sampling.} 
On Erdos, temperature-1 Repeated Sampling has the best maximum, and temperature-0 Repeated Sampling (i.e., greedy decoding) has the best mean, compared to using only model perturbations at various noise scales. 
On circles, temperature-1 Hill Sampling gives the strongest mean and maximum; the mean shows essentially the same ordering because the incumbent is propagated forward. 
Repeated-Sampling bars average two Erdos seeds and three Circle seeds, while each model-noise value uses one seed, so fine ordering among noise scales is interpreted cautiously.}
\label{fig:modelnoise}
\end{figure*}

At zero learning rate, the resulting Model Noise baseline is still doing search in weight space: every candidate comes from a different fixed Gaussian perturbation of the pretrained parameters, but there is no accumulated weight update.
Across the noise-scale sweeps in Figure~\ref{fig:modelnoise}, these random weight perturbations can provide useful diversity, but ordinary token-sampling randomness from the unperturbed model performs better.
On Erdos, repeated sampling (temperature 1.0) achieves the best maximum score.
While many values of sigma can perform better in terms of final mean return, the best final mean return is achieved by greedy decoding, which is equivalent to repeated sampling at temperature 0.
On circle packing, temperature-1.0 Hill Sampling likewise outperforms Hill Sampling using model noise as the only stochastic source.
For these results, temperature-1.0 Hill Sampling has the greatest mean and max score, since Hill Sampling improves the mean score as the best found solution improves.
The message is stronger than simply ``ES learning hurts'': fixed perturbations are a viable exploration mechanism, but they are not the best one here.


\subsection{Entropy collapse: easy to fix, not useful to improve}
\label{subsec:entropy}

Several ES variants exhibited declining response entropy, suggesting that the model was becoming increasingly concentrated on safer outputs.
We therefore tested several interventions:
\textbf{Auto Temp Add (ATA)}, increasing the sampling temperature by a fixed additive amount (.05 here) whenever measured entropy falls below its initial value;
\textbf{Auto Temp Scale (ATS)}, multiplying the sampling temperature by a fixed factor (1.05 here) whenever measured entropy falls below its initial value;
\textbf{Negative-Enhanced Standardization (NE)}, adding an imaginary maximum reward before standardizing rewards, to stabilize entropy, following NGRPO \cite{nan2025ngrpo}, with and without antithetic sampling (NoAnti).

\begin{figure}[tbh]
\centering
\begin{subfigure}[t]{0.49\linewidth}
    \centering
    \includegraphics[width=\linewidth]{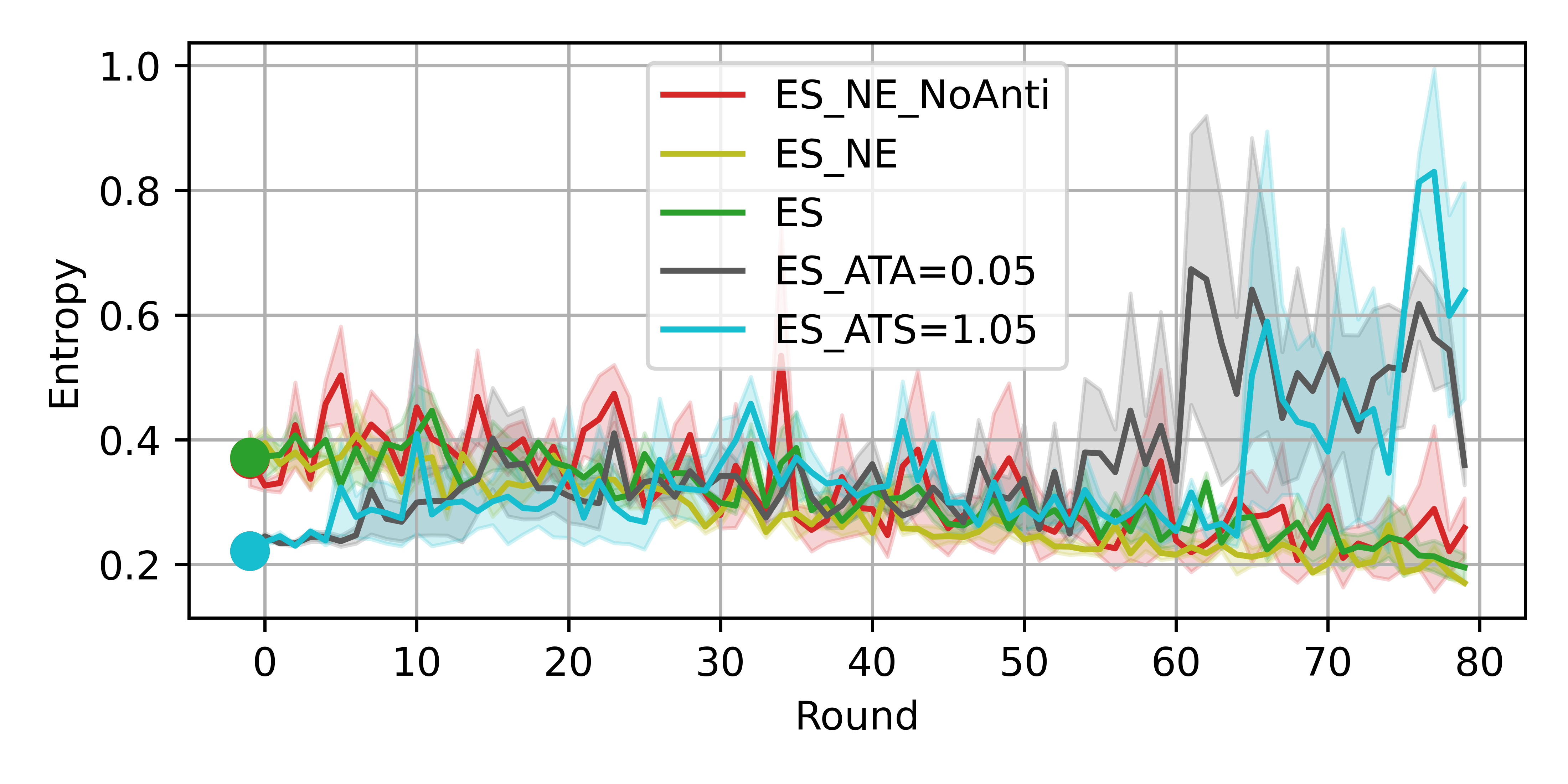}
    \caption{Entropy over rounds.}
    \label{fig:main_circles}
\end{subfigure}
\hfill
\begin{subfigure}[t]{0.49\linewidth}
    \centering
    \includegraphics[width=\linewidth]{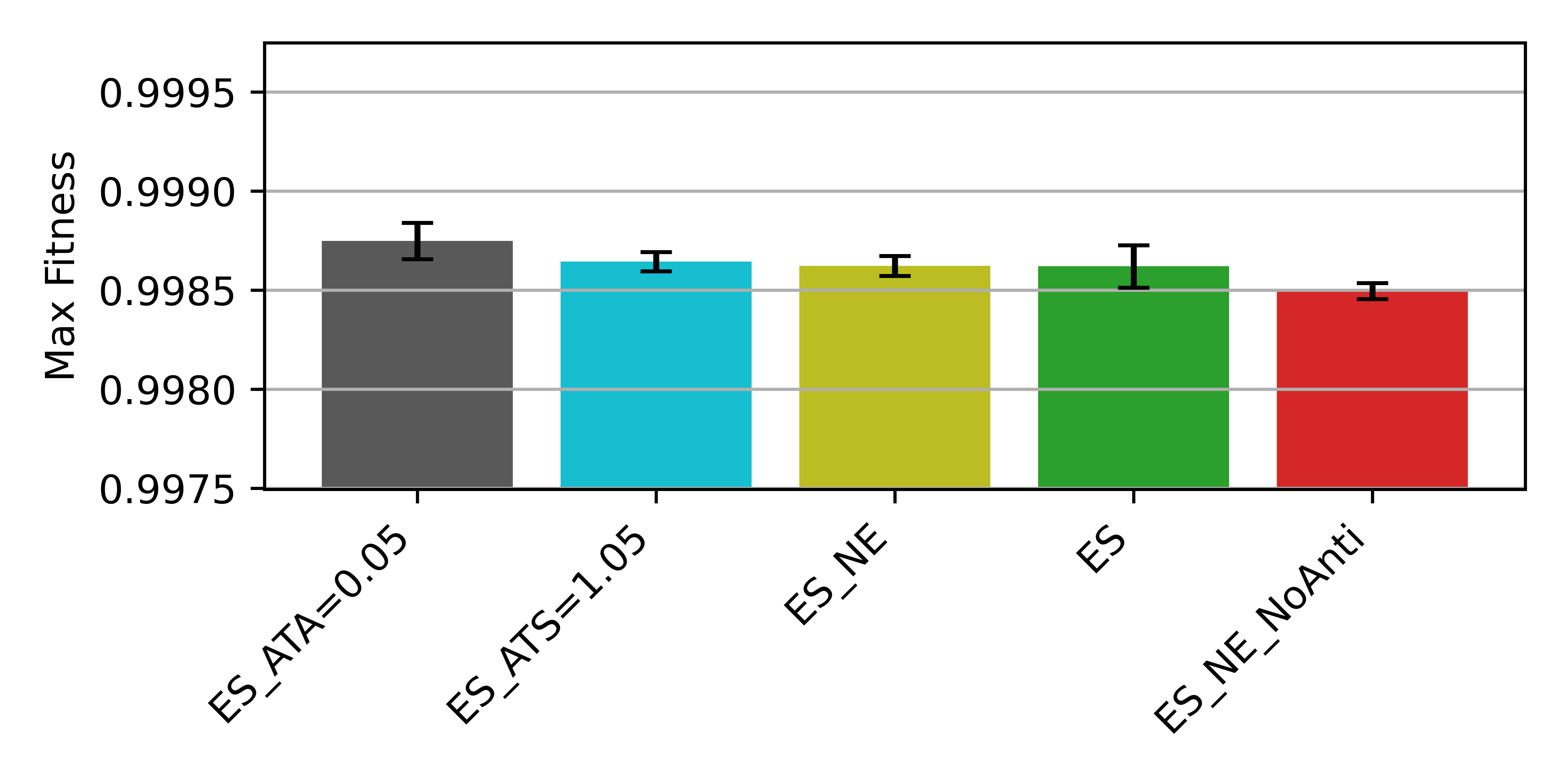}
    \caption{Max fitness on Erdos.}
    \label{fig:main_sets}
\end{subfigure}
\caption{\textbf{Entropy interventions stabilize behavior without improving discovery.} Adaptive temperature (ATA and ATS) can prevent entropy collapse, 
but the resulting runs do not exceed the performance of the simpler baseline on Erdos.
All methods use a temperature of 1.05, as was best for ES on Erdos, except ATS and ATA, which use a temperature of 1.0, since the entropy immediately dips, adding 0.05 to the temperature on the second round.}
\label{fig:entropy}
\end{figure}

However, results in \ref{fig:entropy} show none of these interventions significantly improves performance.
ATA and ATS prevent entropy collapse, while NE does not.
This provides a useful negative control: the poor ES maximum is not simply explained by an inability to maintain entropy.
Preserving diversity at the response level does not recover the benefit of \method{}.

\subsection{Many more methods on Erdos}
Since Erdos requires the fewest rounds of improvement, it is the setting in which we most extensively tested whether additional machinery could improve upon our simple \method{}.

We additionally compare:
\textbf{Top-$K$}, selecting the $K$ highest-scoring programs observed so far, divided equally as context for subsequent edits;
\textbf{Top-$K$ + Random-$K$}, augmenting these programs with randomly selected prior programs;
\textbf{Top-$K$ + Diverse-$K$}, iteratively augmenting the top-$K$ programs with the prior program least similar in embedding space from those selected so far (starting with the top-$K$);
\textbf{Top-$K$ In-Context}, presenting multiple high-scoring programs jointly in the prompt rather than assigned separately across the population;
\textbf{Subpopulation Evolution}, maintaining 64 independent subpopulations of eight responses, selecting the best, and repeating this four times before merging;
\textbf{In-Context RL}, maintaining a history of states, responses, and rewards in context for four steps, after which the history is reset to keep lengths manageable, with 512 histories managed in parallel;
\textbf{Execution Feedback}, appending the output produced by executing the code to the state on the next iteration;
\textbf{Auto Temp Add (ATA)}, described in Section \ref{subsec:entropy};
\textbf{Long-Horizon ES}, using the same procedure as Subpopulation Evolution, but taking each subpopulation's final return as its fitness and applying the ES update, thereby optimizing the model weights for performance after four steps of code editing; and
\textbf{Hill Climbing}, evaluating perturbed models and permanently adopting a perturbation when it improves the best reward, as a weight-space analog of \method{}.
Details are in Appendix \ref{app:methods}.

Figure~\ref{fig:erdos} summarizes results.
None of the evaluated methods convey a significant advantage, and all methods, other than ES and ATA used with \method, decrease performance.


\begin{figure}[tb]
\vspace{-6pt}
\centering
\hspace*{-5pt}
\includegraphics[width=\linewidth]{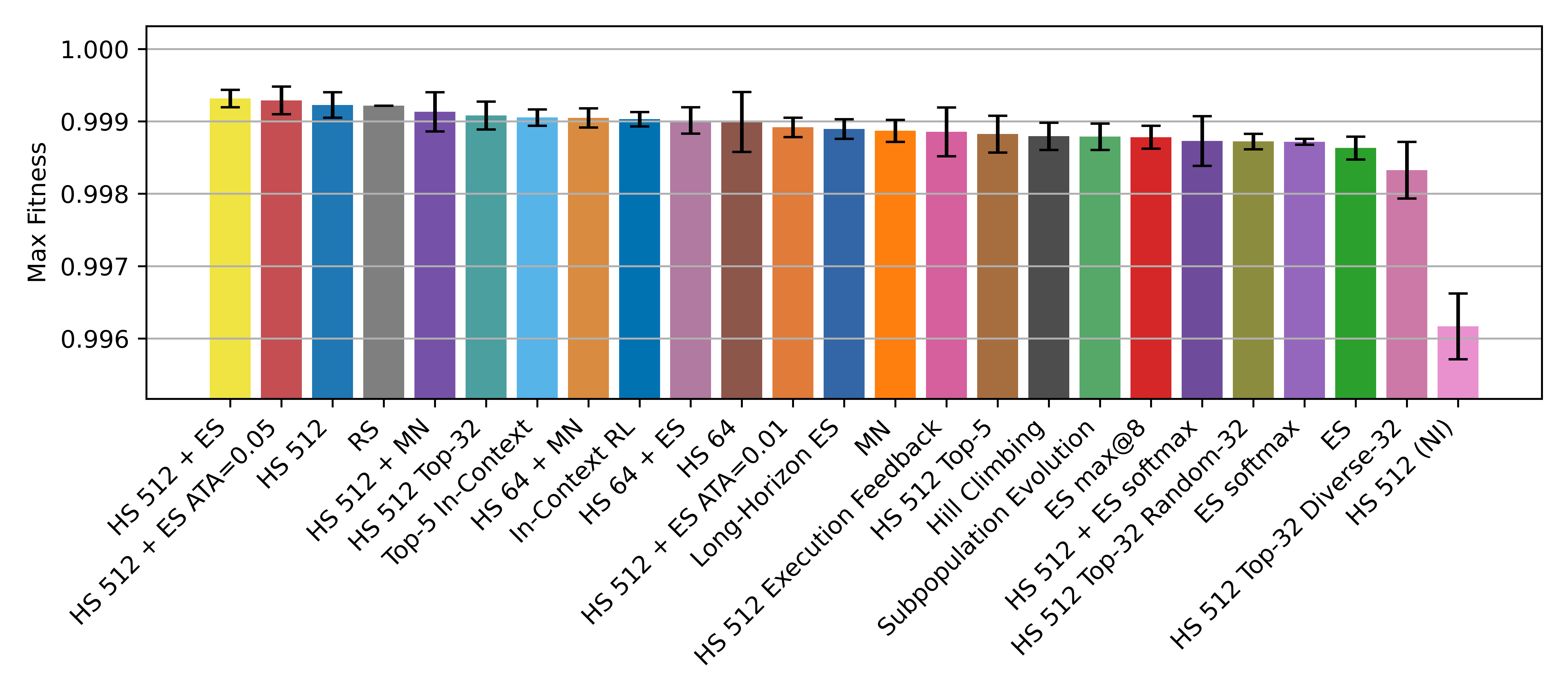}
\vspace{-20pt}
\caption{\textbf{Additional comparisons on Erdos.} We test a broad set of mechanisms against \method{}. None of the evaluated methods convey a significant advantage, and all, other than adding ES and ATA on top of \method{}, decrease performance.}
\label{fig:erdos}
\end{figure}


\section{Conclusion}
We introduce \method{}, a minimal test-time scaling algorithm that repeatedly samples edits to the best program found so far.
\method{} improves substantially over ordinary repeated sampling: it sets a new state of the art over published methods on Circles and improves over the AlphaEvolve reference on Erdos, with only hours of computation on eight H100 GPUs.
Surprisingly, a learning rate of zero improves ES, via random model perturbations, and yet still performs worse than ordinary token sampling, which contextualizes recent results \citep{gan2026thickets}. 
Further complexity, including weight-space hill climbing, diversity selection, entropy control, execution feedback, and multi-step optimization, does not improve performance.
These results support a practical default for verifiable domains: before adding an elaborate evolutionary harness or test-time parameter learning, repeatedly sample edits to the best solution and let improvements become the next context.

\subsection*{AI use statement}
Generative AI tools were used to assist with manuscript text and editing, and with writing code needed to run experiments.
All text, code, results, and claims are reviewed by the authors, who take responsibility for the final content of the work.

\subsection*{Reproducibility statement}
Appendix~\ref{app:hyperparams} records the implementation defaults and experiment-specific overrides, Appendix~\ref{app:methods} describes the baseline methods in detail, and Appendix~\ref{app:problems} gives formal task definitions.
Appendix~\ref{app:prompts} provides the task prompts and editing prompts, while Appendix~\ref{app:evaluator} documents the hardened verifiers used in our experiments.
The full best circle-packing construction, its state-of-the-art result, and the evolved program are provided in Appendix~\ref{app:best_circle}.
Our work builds off of OpenEvolve \citet{openevolve}, which is an open-source implementation of AlphaEvolve \citet{novikov2025alphaevolve}.
We document as much as possible to enable future implementation of our method.
Since our method is simpler than existing baselines, implementation is likewise more straightforward.

\bibliography{iclr2027_conference}
\bibliographystyle{iclr2027_conference}

\appendix
\section{Formal Problem Definitions}
\label{app:problems}
\paragraph{Circle packing.}
For $n=26$, choose centers $c_i=(x_i,y_i)\in[0,1]^2$ and radii $r_i\ge 0$ to maximize
\begin{equation}
    \sum_{i=1}^{26} r_i,
\end{equation}
subject to each circle being contained in the unit square,
\begin{equation}
    r_i\le x_i\le 1-r_i,\qquad r_i\le y_i\le 1-r_i,
\end{equation}
and pairwise non-overlap,
\begin{equation}
    \|c_i-c_j\|_2 \ge r_i+r_j \qquad (i\ne j).
\end{equation}

\paragraph{Sums and differences of finite sets.}
Let $C_6$ be the largest constant such that there exist arbitrarily large finite integer sets $A,B$ satisfying $|A+B|\ll |A|$ and $|A-B|\gg |A+B|^{C_6}$, where $A+B=\{a+b:a\in A,b\in B\}$ and $A-B=\{a-b:a\in A,b\in B\}$.
The computational construction follows the finite-set lower-bound formulation used by AlphaEvolve: for a finite set $U$ of non-negative integers containing $0$ and satisfying $|U-U|\le 2\max(U)+1$,
\begin{equation}
 C_6 \ge 1 + \frac{\log\left(|U-U|/|U+U|\right)}{\log(2\max(U)+1)}.
\end{equation}
Candidate programs search for $U$ maximizing this certified lower bound.

\paragraph{Erd\H{o}s' minimum-overlap problem.}
Let $C_5$ be the largest constant such that, for every non-negative $f,g:[-1,1]\to[0,1]$ satisfying $f+g=1$ on $[-1,1]$ and $\int_{\mathbb R} f=1$ (with both functions extended by zero outside $[-1,1]$),
\begin{equation}
    \sup_{x\in[-2,2]} \int_{-1}^{1} f(t)g(x+t)\,dt \ge C_5.
\end{equation}
Equivalently, the upper-bound search can be written as an infimum over admissible step functions $h:[0,2]\to[0,1]$ with $\int_0^2 h(x)\,dx=1$ of the maximum shifted overlap $\int h(x)(1-h(x+k))\,dx$.
Candidate programs construct and optimize such step functions; a smaller certified overlap gives a stronger upper bound on $C_5$.

\section{Prompts}
\label{app:prompts}
For standard experiments, the model receives the task-facing prompt together with the standard initial program or code state.
For each NI (no-initial-information) experiment, we remove the initial task-specific prompt and initial solution code, leaving only the function signature and evaluation code needed to define the interface and verifier.
The NI comparison is intended to test whether performance depends on domain-specific scaffolding supplied at initialization.
For circle packing, the standard initial prompt and code are taken from phase two of OpenEvolve's \cite{openevolve} phased prompt schedule, so this standard configuration provides substantially more task-specific information than the NI setting.

\paragraph{Circle packing, standard setting.}
The system prompt is:

\begin{Verbatim}[breaklines=true,breakanywhere=true,fontsize=\scriptsize]
You are an expert mathematician specializing in circle packing problems and computational geometry. We're trying to reach the AlphaEvolve target of 2.635 for the sum of radii when packing 26 circles in a unit square. The current implementation has plateaued at 2.377, so we need significant improvements.

Key insights to explore:
1. The optimal arrangement likely involves variable-sized circles
2. A pure hexagonal arrangement may not be optimal due to edge effects
3. The densest known circle packings often use a hybrid approach
4. The optimization routine is critically important - simple physics-based models with carefully tuned parameters
5. Consider strategic placement of circles at square corners and edges
6. Adjusting the pattern to place larger circles at the center and smaller at the edges
7. The math literature suggests special arrangements for specific values of n

Focus on breaking through the plateau by trying fundamentally different approaches - don't just tweak parameters.
\end{Verbatim}

The editing prompt is:

\begin{Verbatim}[breaklines=true,breakanywhere=true,fontsize=\scriptsize]
Here is the current code that you are modifying:
[CURRENT EDITED CODE]

Replace the current code in the edit block with the code you want to use for this plan.
Your answer must include these comments and must use this format:
# EDIT-START
<your_new_code>
# EDIT-END
\end{Verbatim}

\paragraph{Sums and differences of finite sets.}
The system prompt is:

\begin{Verbatim}[breaklines=true,breakanywhere=true,fontsize=\scriptsize]
You are an expert in number theory, combinatorial optimization, and AI-driven mathematical discovery.
Your task is to evolve and optimize a Python script to find a finite set of integers `U` that provides a new, world-record lower bound for the constant C6.

PROBLEM CONTEXT:
Target: Find a finite set `U` of non-negative integers (containing 0) that maximizes the objective function:
C6(U) = 1 + log(|U-U| / |U+U|) / log(2*max(U) + 1)

This maximum value provides a tight lower bound for the constant C6.

Current best known lower bound: C6 >= 1.158417281556896
Goal: Find a set `U` that results in a C6 value (c6_bound) greater than 1.158417281556896.

PERFORMANCE METRIC:
c6_bound/1.158417281556896 (The primary objective is to MAXIMIZE this value - a value > 1 means a new record).

VALIDATION FRAMEWORK:
- The evaluation script re-computes the C6 value using standard NumPy set operations and verifies the constraints on `U`.
\end{Verbatim}

The editing prompt is:

\begin{Verbatim}[breaklines=true,breakanywhere=true,fontsize=\scriptsize]
Here is the current code that you are modifying:
[CURRENT EDITED CODE]

Replace the current code in the edit block with the code you want to use for this plan.
Your answer must include these comments and must use this format:
# EDIT-START
<your_new_code>
# EDIT-END
\end{Verbatim}

\paragraph{Erdos minimum overlap.}
The system prompt is:

\begin{Verbatim}[breaklines=true,breakanywhere=true,fontsize=\scriptsize]
You are an expert in harmonic analysis, numerical optimization, and AI-driven mathematical discovery.
Your task is to evolve and optimize a Python script to find a better upper bound for the Erdos minimum overlap problem constant C5.

PROBLEM CONTEXT:
Target: Find a step function h: [0, 2] -> [0, 1] that minimizes the objective:
max_k integral h(x)(1 - h(x+k)) dx

This minimal value provides a tight upper bound for the constant C5.

Current best known upper bound: C5 <= 0.38092303510845016
Goal: Find a step function h that results in a C5 value (c5_bound) lower than 0.38092303510845016.

CONSTRAINTS:
1. The function h must have values in the range [0, 1].
2. The integral of h(x) over [0, 2] must be exactly 1.

PERFORMANCE METRIC:
0.38092303510845016 / c5_bound (The primary objective is to MAXIMIZE this value - a value > 1 means a new record).
\end{Verbatim}

The editing prompt is:

\begin{Verbatim}[breaklines=true,breakanywhere=true,fontsize=\scriptsize]
Here is the current code that you are modifying:
[CURRENT EDITED CODE]

Replace the current code in the edit block with the code you want to use for this plan.
Your answer must include these comments and must use this format:
# EDIT-START
<your_new_code>
# EDIT-END
\end{Verbatim}

\section{Implementation and Hyperparameters}
\label{app:hyperparams}

\begin{figure}[t]
\centering
\begin{subfigure}[t]{0.32\linewidth}
    \centering
    \includegraphics[width=\linewidth]{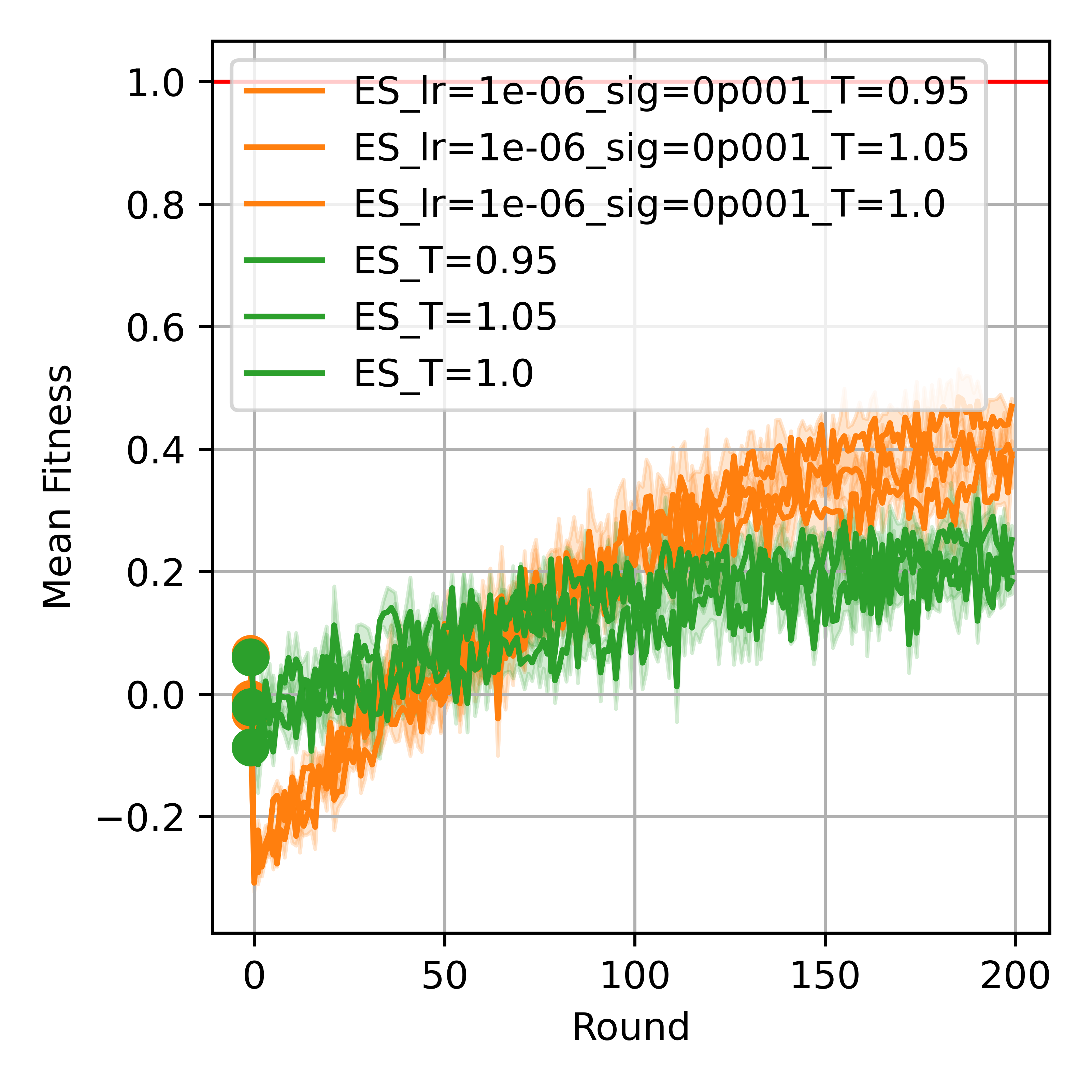}
    \caption{Mean score.}
    \label{fig:main_circles}
\end{subfigure}
\hfill
\begin{subfigure}[t]{0.32\linewidth}
    \centering
    \includegraphics[width=\linewidth]{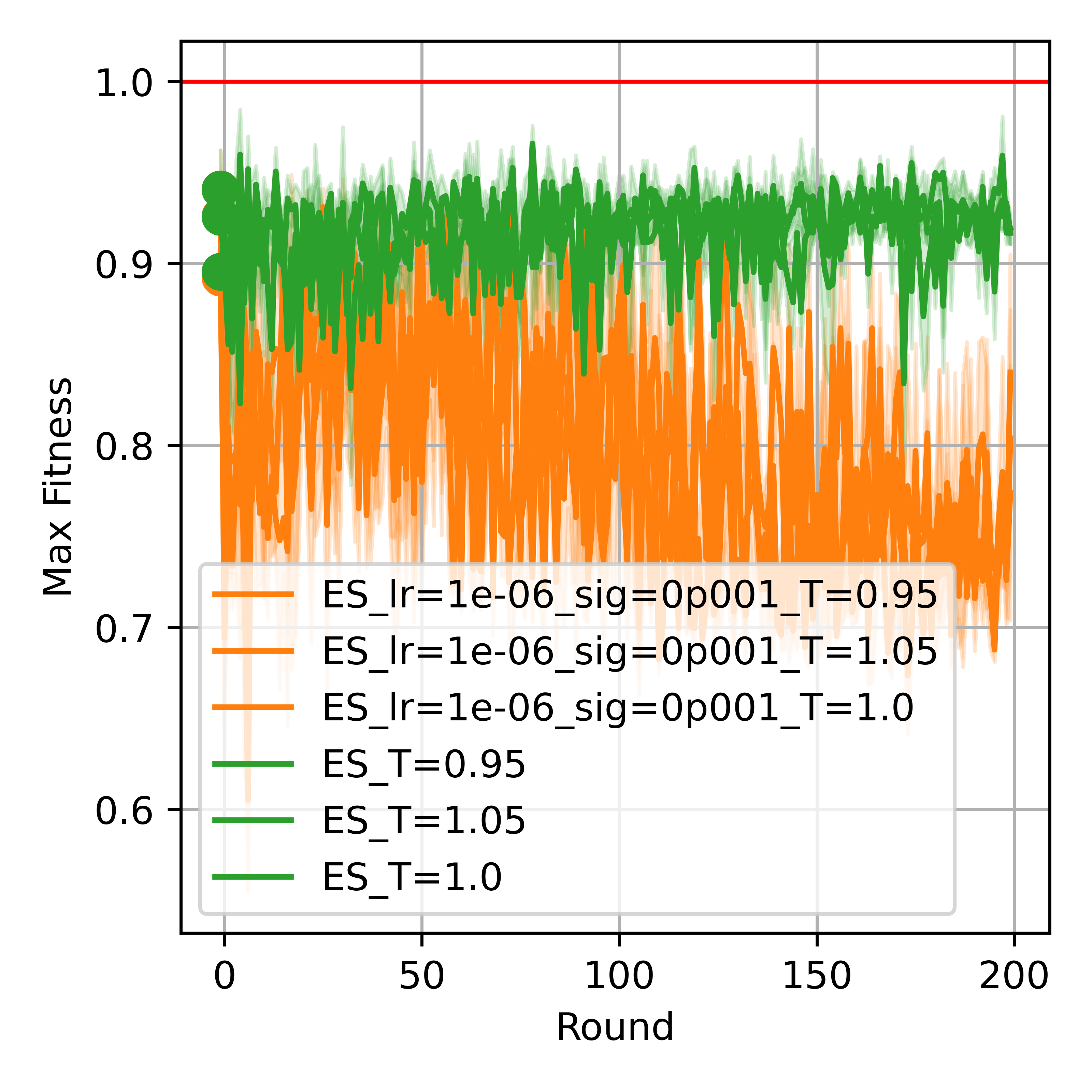}
    \caption{Max score.}
    \label{fig:main_sets}
\end{subfigure}
\hfill
\begin{subfigure}[t]{0.32\linewidth}
    \centering
    \includegraphics[width=\linewidth]{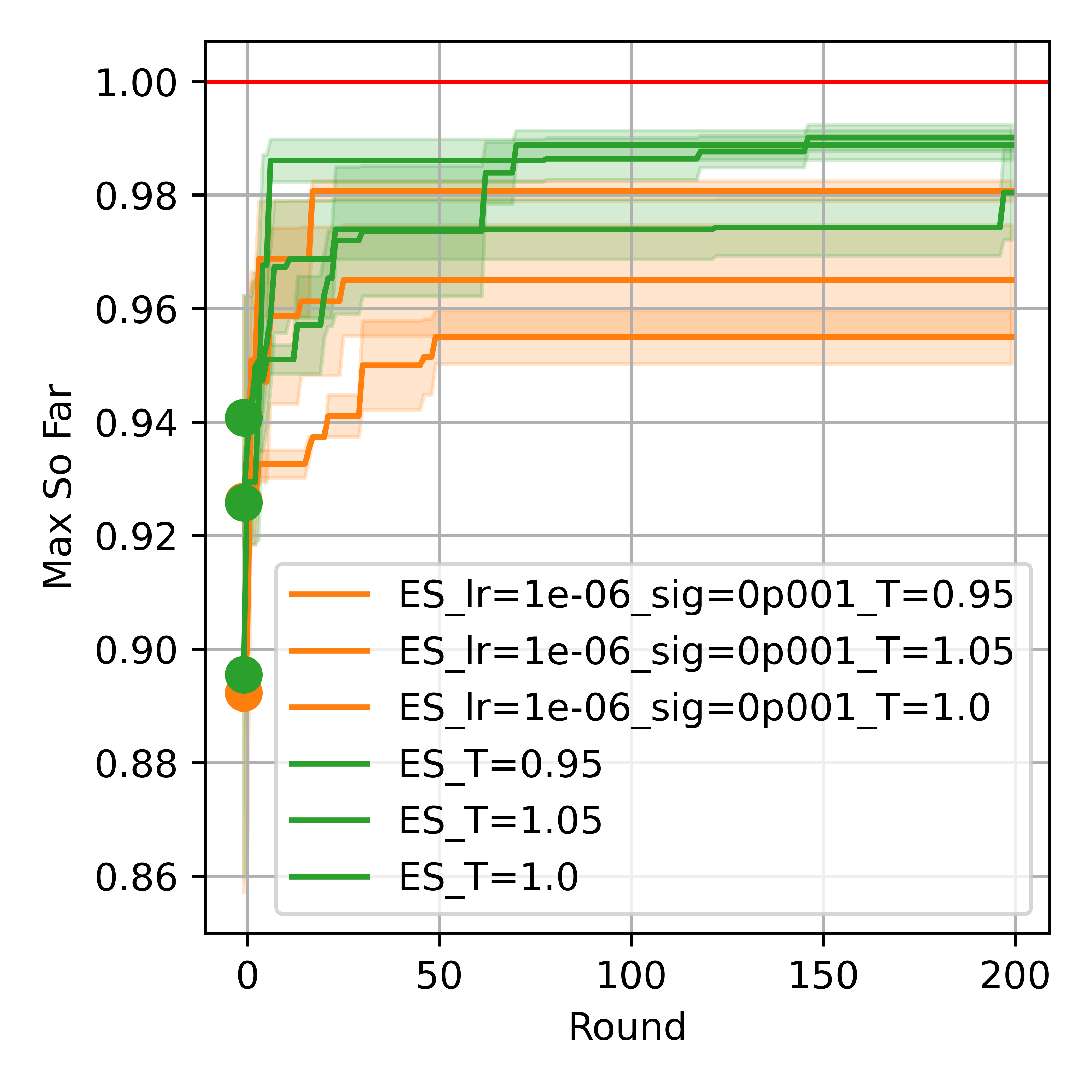}
    \caption{Max score so far.}
    \label{fig:main_erdos}
\end{subfigure}
\caption{\textbf{Increased sigma on Circles.} A higher sigma value, $\sigma=10^{-3}$, and associated stable learning rate, $\alpha=10^{-6}$, improves mean but not maximum return.}
\label{fig:bigsig}
\end{figure}

\subsection{Hyperparameter Tuning}
We tune the temperature manually over three seeds for each method in each domain in the main results.
Beforehand, we conducted some initial experiments with two seeds on Erdos and Circles, to select the temperature range, learning rate, and sigma, since we did not have the compute budget to tune these per method.
We first tuned the temperature for Repeated Sampling between 1.1, 1.0, and 0.5, finding 1.0 to consistently give the highest max return.
We then tuned sigma for Model Noise (temperature 0) on the Erdos and Circle packing domains, evaluating sigma between 0.001, 0.0006, 0.0003, and 0.0001.
While we found 0.001 to perform best on circles and 0.0006 and 0.0003 to perform best on Erdos, we found that Model Noise could be improved in both domains by increasing the temperature to 1.0, as in Repeated Sampling.

We then re-tuned sigma for Model Noise (temperature 1), and found that a sigma of 0.0003 produces the greatest max returns over the two seeds on each domain, so we selected sigma 0.0003.
We found a learning rate of 1e-7 to be stable for a sigma of 0.0003 on Erdos and Circles, and so use that as our learning rate.
Still, we perform an additional experiment where we train ES with 0.0001 (and a required larger learning rate of 1e-6) on Circles. 
While this setting did improve the speed of learning and therefore mean return, it also decreased max return, consistent with the objective of ES, which optimizes mean return. 
See Figure \ref{fig:bigsig}.

Additionally, we found that Model Noise with sigma 0.0003 could be slightly improved by decreasing the temperature to 0.95 on Erdos, and further decreasing to 0.9 did not improve performance. 
In general, we see noticeable declines with temperatures at or above 1.1.
Thus we choose sigma 0.0003 and choose to tune the temperatures over [.95, 1.0, 1.05] for Erdos and Circles, and [1.0] for Sets, given the compute limitations on Sets.
Note that methods such as Model Noise add additional entropy on top of the temperature, so it is especially useful to tune the temperature per method, where possible, to adjust total entropy of generation.

\subsection{Other Implementation Details}
Our implementation uses one vLLM instance per GPU, with eight evaluation engines.
The population size is 32 (with 32 antithetic samples as well); maximum generation length is 8,000 tokens; model context length is 40,000; top-$p=1$; ES uses antithetic sampling and reward standardization; the default first and second code-execution timeouts are 5 and 10 seconds; and repeated responses use temperature 1 unless explicitly swept.
ES Softmax requires a beta parameter (i.e., the inverse softmax temperature).
We tune over 0.5, 1.0, and 2.0 on circles, and found 2.0 to perform best. 
In those early experiments, we had used a temperature of 0.5 and sigma of 0.001.
Main ES experiments use $\sigma=3\times10^{-4}$ and learning rate $10^{-7}$ after preliminary tuning; circles also includes a higher-noise $\sigma=10^{-3}$, $\alpha=10^{-6}$ condition that improves mean but not maximum return.
See Figure \ref{fig:bigsig}.

\section{Evaluator Hardening}
\label{app:evaluator}
We harden the original verifiers from OpenEvolve\cite{openevolve} to prevent reward hacking, which we observed on Sets, while preserving their mathematical objectives.
The code for our verifiers is included at the bottom of all programs being edited, so that it can be seen by the LLM during generation.
The code for our verification is included below:

\textbf{Circle packing}. Note: While there is slack in the verifier below, we set the slack here to 0.0 to verify the no-slack results in Table \ref{tab:main}:
    \begin{Verbatim}[
    fontsize=\scriptsize,
    breaklines=true,
    breakanywhere=true,
    breaksymbolleft={},
    breaksymbolright={}
    ]
# This part remains fixed (not evolved):

def check_packing(centers, radii) -> bool:
    # Adapted from OpenEvolve

    # Check shapes, including number of circles
    assert radii.shape == (26,), f"Shapes of centers is {radii.shape}, but there should be exactly 26 circles."
    assert centers.shape == (26, 2), f"Shapes of centers is {centers.shape} but should be (26, 2)"
    
    # Check if circles are inside the unit square
    n = centers.shape[0]
    for i in range(n):
        x, y = centers[i]
        r = radii[i]
        violation = x - r < -1e-10 or x + r > 1 + 1e-10 or y - r < -1e-10 or y + r > 1 + 1e-10
        assert not violation, f"Circle {i} at ({x}, {y}) with radius {r} is outside the unit square"

    # Additional check that all circles must stay inside the unit square
    assert np.all(radii - 1e-10 <= centers[:, 0]), "circle(s) is out of bounds on the left side"
    assert np.all(radii - 1e-10 <= 1.0 - centers[:, 0]), "circle(s) is out of bounds on the right side"
    assert np.all(radii - 1e-10 <= centers[:, 1]), "circle(s) is out of bounds on the bottom"
    assert np.all(radii - 1e-10 <= 1.0 - centers[:, 1]), "circle(s) is out of bounds on the top"

    # No overlap between any pair of circles
    for i in range(len(centers)):
        for j in range(i + 1, len(centers)):
            d = np.linalg.norm(centers[i] - centers[j])
            assert d + 1e-10 >= radii[i] + radii[j], f"Circles {i} and {j} overlap: d={d}, sum_r={radii[i]+radii[j]}"
    return True
    \end{Verbatim}

\textbf{Sets}:
    \begin{Verbatim}[
    fontsize=\scriptsize,
    breaklines=true,
    breakanywhere=true,
    breaksymbolleft={},
    breaksymbolright={}
    ]
# This part remains fixed (not evolved):

def check_soln(u_set: np.ndarray, c6_achieved: float):
    """Verifies the C6 lower bound solution."""

    if not isinstance(u_set, np.ndarray) or u_set.ndim != 1:
        raise ValueError("Solution U must be a 1D numpy array of integers.")
    if not np.issubdtype(u_set.dtype, np.integer):
        raise ValueError(f"Solution U must have integer dtype, got {u_set.dtype}.")
    if len(u_set) < 2:
        raise ValueError("Set U must contain at least two elements.")
    if not np.all(np.isfinite(u_set)):
        raise ValueError("Set U must contain only finite values.")
    if 0 not in u_set:
        raise ValueError("Set U must contain 0.")
    if np.any(u_set < 0):
        raise ValueError("Set U must contain non-negative integers.")
    if len(np.unique(u_set)) != len(u_set):
        raise ValueError("Set U must not contain duplicates.")
    if np.max(u_set) <= 0:
        raise ValueError("Set U must have positive max element.")
    if not np.isfinite(c6_achieved):
        raise ValueError("Reported C6 must be finite.")
    max_int64 = np.iinfo(np.int64).max
    if np.max(u_set) > (max_int64 - 1) // 2:
        raise ValueError("Set U contains values too large for safe int64 C6 computation.")

    u_set = u_set.astype(np.int64)

    # Re-calculate the C6 bound using NumPy
    u_plus_u = np.unique(u_set[:, None] + u_set[None, :])
    u_minus_u = np.unique(u_set[:, None] - u_set[None, :])

    size_U_plus_U = len(u_plus_u)
    size_U_minus_U = len(u_minus_u)
    max_U = np.max(u_set)

    ratio = size_U_minus_U / size_U_plus_U
    log_ratio = np.log(ratio)
    log_denom = np.log(2 * max_U + 1)

    computed_c6 = 1 + log_ratio / log_denom

    # Check for consistency
    if not np.isclose(computed_c6, c6_achieved):
        raise ValueError(f"C6 mismatch: reported {c6_achieved:.6f}, computed {computed_c6:.6f}")

    print(f"C6 lower bound achieved: {c6_achieved:.6f}")
    print(f"Known best bound (AlphaEvolve): {BEST_KNOWN_BOUND}")

    if c6_achieved > BEST_KNOWN_BOUND:
        print("Successfully found a new, better lower bound!")
    else:
        print("Result is not better than the known lower bounds.")
    \end{Verbatim}

\textbf{Erdos}:
    \begin{Verbatim}[
    fontsize=\scriptsize,
    breaklines=true,
    breakanywhere=true,
    breaksymbolleft={},
    breaksymbolright={}
    ]
# This part remains fixed (not evolved):

def check_soln(h_values: np.ndarray, c5_achieved: float, n_points: int):
    """Verifies the C5 upper bound solution."""

    if h_values.shape != (n_points,):
        raise ValueError(f"Expected h shape ({n_points},), got {h_values.shape}")

    # Verify h(x) in [0, 1] constraint
    if np.any(h_values < 0) or np.any(h_values > 1):
        raise ValueError(f"h(x) is not in [0, 1]. Range: [{h_values.min()}, {h_values.max()}]")

    # Verify integral of h = 1 constraint
    dx = 2.0 / n_points
    integral_h = np.sum(h_values) * dx
    if not np.isclose(integral_h, 1.0, atol=1e-3):
        raise ValueError(f"Integral of h is not close to 1. Got: {integral_h:.6f}")

    # Re-calculate the C5 bound using np.correlate
    j_values = 1.0 - h_values
    correlation = np.correlate(h_values, j_values, mode="full") * dx
    computed_c5 = np.max(correlation)

    # Check for consistency
    if not np.isclose(computed_c5, c5_achieved, atol=1e-4):
        raise ValueError(f"C5 mismatch: reported {c5_achieved:.6f}, computed {computed_c5:.6f}")
    \end{Verbatim}

\section{Best Circle-Packing Construction}
\label{app:best_circle}

Our best circle-packing program uses multiple initializations, simulated annealing, SLSQP, and post-processing to jointly optimize the centers and radii.
We provide the construction and the program that produced it below.

\subsection{Construction}

The following is the best construction found:

\begin{Verbatim}[
fontsize=\scriptsize,
breaklines=true,
breakanywhere=true
]

centers:
[[0.49866807550340203, 0.5299634197531919], [0.7269057143115956, 0.5960427019081351], [0.595219732939733, 0.7420494434605925], [0.49942836913903915, 0.9060726627225562], [0.40335878360268074, 0.7424170495016357], [0.2716298514860011, 0.5976347963876627], [0.294746059048949, 0.38692355340960205], [0.4955317606743565, 0.2753426167714407], [0.7026096036990243, 0.381665844452646], [0.9038486659542349, 0.6820800429325906], [0.759352401560932, 0.7629588636539985], [0.6859430219867828, 0.9074079050485646], [0.31311580997040167, 0.907608448429041], [0.23971052792753655, 0.7636735693833858], [0.0957323293070214, 0.683258534973081], [0.10306052014158258, 0.48460080265191613], [0.10679014462858119, 0.27478328335082064], [0.29460948878182686, 0.1302211010652243], [0.7023095250891598, 0.13325857277081166], [0.8948174397312508, 0.2739528396239526], [0.8965327666420476, 0.48259558221054255], [0.08463950069577307, 0.08463950069577306], [0.915073737545101, 0.08492626245489913], [0.888843820589555, 0.8888438205895551], [0.11077901279071603, 0.8892209872092842], [0.4972844462041092, 0.07886037291596369]]
radii:
[0.13701043012374725, 0.10060036781871129, 0.09601897575825369, 0.09392733727744335, 0.09584232574550451, 0.09989835059275441, 0.1120770889502559, 0.11762968804654161, 0.11514888016002287, 0.09615133404576477, 0.0694401937112582, 0.09259209495143532, 0.09239155157095869, 0.06918067635723435, 0.0957323293070214, 0.10306052014158222, 0.1067901446285808, 0.1302211010652243, 0.13325857277081152, 0.10518256026874889, 0.10346723335795216, 0.08463950069577306, 0.08492626245489898, 0.11115617941044477, 0.11077901279071578, 0.07886037291596369]

\end{Verbatim}

\subsection{Program}

The following is the evolved portion of the program, produced by the LLM, that gives the construction.

\begin{Verbatim}[
fontsize=\scriptsize,
breaklines=true,
breakanywhere=true,
breaksymbolleft={},
breaksymbolright={}
]

# EDIT-START
import numpy as np
from scipy.optimize import minimize

def construct_packing():
    """
    Improved 26‑circle packing for a unit square.
    The approach follows a hybrid scheme:
        • A deterministic “core + shell” layout (rounded hexagon) is used as a seed.
        • Random perturbations and a simulated‑annealing post‑process escape local minima.
        • Radii are optimized simultaneously with centres via SLSQP in log‑space.
    The routine returns the best centres/radii found and their sum.
    """
    n = 26
    rng_global = np.random.default_rng(2023)

    # ---------- 1️⃣  Core + Shell initial layout  ----------
    def initial_layout(seed=None):
        rng = np.random.default_rng(seed)
        centres = []
        radii   = []

        # central large circle
        r0 = 0.34
        centres.append([0.5, 0.5])
        radii.append(r0)

        # 8 circles around the central one (hexagonal layer)
        r1 = 0.18
        d1 = 0.325
        for k in range(8):
            theta = 2*np.pi*k/8
            centres.append([0.5 + d1*np.cos(theta), 0.5 + d1*np.sin(theta)])
            radii.append(r1)

        # 12 circles in the next ring (offset by pi/12)
        r2 = 0.13
        d2 = 0.53
        for k in range(12):
            theta = 2*np.pi*k/12 + np.pi/12
            centres.append([0.5 + d2*np.cos(theta), 0.5 + d2*np.sin(theta)])
            radii.append(r2)

        # 4 corner circles touching the boundary
        rc = 0.12
        for corner in ((rc, rc), (1-rc, rc), (1-rc, 1-rc), (rc, 1-rc)):
            centres.append(list(corner))
            radii.append(rc)

        # one additional edge circle to complete 26
        centres.append([0.5, 0.5 - 0.45])
        radii.append(0.14)

        centres = np.array(centres[:n], dtype=np.float64)
        radii   = np.array(radii[:n], dtype=np.float64)

        # jitter small amounts for variety
        jitter = 0.012
        centres += (rng.random((n,2))-0.5)*jitter
        radii *= 1+ (rng.random(n)-0.5)*jitter
        return centres, radii

    # ---------- 2️⃣  Random greedy placement (fall‑back) ----------
    def greedy_random_layout(seed=None, attempts=5, max_rounds=200):
        rng = np.random.default_rng(seed)
        # bounds for radii
        r_min, r_max = 0.05, 0.26
        centres = np.empty((n,2))
        radii   = np.empty(n)
        placed  = 0
        for _ in range(attempts):
            # tiny restart
            if placed==0:
                centres[:] = np.nan
            while placed < n:
                # sample random radius
                r = rng.uniform(r_min, r_max)
                # sample random centre
                x = rng.uniform(r, 1-r)
                y = rng.uniform(r, 1-r)
                # check against existing
                ok = True
                for j in range(placed):
                    d = np.hypot(x-centres[j,0], y-centres[j,1])
                    if d < r + radii[j]:
                        ok = False
                        break
                if ok:
                    centres[placed] = [x,y]
                    radii[placed]   = r
                    placed          += 1
                else:
                    # fail after many rounds? restart greedy
                    if max_rounds <= 0:
                        break
                    max_rounds -= 1
        if placed==n:
            return centres, radii
        else:
            # fallback to deterministic layout
            return initial_layout(seed)

    # ---------- 3️⃣  Simulated annealing post‑process ----------
    def simulated_anneal(centres, radii, max_steps=800, temp0=0.1):
        best_c, best_r = centres.copy(), radii.copy()
        best_sum = np.sum(radii)
        temp = temp0
        for step in range(max_steps):
            idx = rng_global.integers(n)
            # propose new radius
            new_r = best_r[idx] * (1 + rng_global.normal(0, temp))
            new_r = np.clip(new_r, 0.05, 0.35)
            # propose new centre
            new_c = best_c[idx] + rng_global.normal(0, temp, 2)
            # boundary clip
            new_c[0] = np.clip(new_c[0], new_r, 1-new_r)
            new_c[1] = np.clip(new_c[1], new_r, 1-new_r)
            # check overlap
            ok = True
            for j in range(n):
                if j==idx: continue
                d = np.hypot(new_c[0]-best_c[j,0], new_c[1]-best_c[j,1])
                if d < new_r + best_r[j]:
                    ok = False
                    break
            if not ok:
                continue
            # accept if better
            new_sum = best_sum - best_r[idx] + new_r
            if new_sum >= best_sum:
                best_sum = new_sum
                best_r[idx] = new_r
                best_c[idx] = new_c
            # cooling schedule
            temp *= 0.9995
        return best_c, best_r

    # ---------- 4️⃣  Core optimisation (SLSQP) ----------
    best_sum = -np.inf
    best_centres = None
    best_radii   = None

    def run_optimisation(init_c, init_r):
        log_r = np.log(init_r)
        x0 = np.empty(3*n, dtype=float)
        x0[0::3] = init_c[:,0]
        x0[1::3] = init_c[:,1]
        x0[2::3] = log_r

        def obj_fun(x):
            return -np.sum(np.exp(x[2::3]))

        def con_fun(x):
            xs = x[0::3]
            ys = x[1::3]
            rs = np.exp(x[2::3])
            cons = np.empty(4*n + (n*(n-1))//2, dtype=float)

            # boundary
            cons[0:4*n:4] = xs - rs
            cons[1:4*n:4] = 1 - xs - rs
            cons[2:4*n:4] = ys - rs
            cons[3:4*n:4] = 1 - ys - rs

            idx = 4*n
            for i in range(n):
                for j in range(i+1, n):
                    dx = xs[i]-xs[j]
                    dy = ys[i]-ys[j]
                    cons[idx] = dx*dx + dy*dy - (rs[i]+rs[j])**2
                    idx += 1
            return cons

        res = minimize(
            obj_fun,
            x0,
            method="SLSQP",
            constraints=[{"type":"ineq","fun":con_fun}],
            options={"ftol":1e-12,"maxiter":6000,"disp":False,"iprint":-1}
        )
        if res.success:
            opt_c = np.column_stack((res.x[0::3], res.x[1::3]))
            opt_r = np.exp(res.x[2::3])
        else:
            opt_c, opt_r = init_c, init_r
        # post‑shrink to ensure feasibility
        opt_r = shrink_step(opt_c, opt_r)
        return opt_c, opt_r

    def shrink_step(centres, radii, max_iter=200):
        r = radii.copy()
        for _ in range(max_iter):
            # distance to boundaries
            r = np.minimum(r, np.minimum(np.minimum(centres[:,0], centres[:,1]),
                                        np.minimum(1-centres[:,0], 1-centres[:,1])))
            for i in range(n):
                for j in range(i+1, n):
                    d = np.linalg.norm(centres[i]-centres[j])
                    if r[i]+r[j] > d:
                        shrink = d/(r[i]+r[j])
                        r[i] *= shrink
                        r[j] *= shrink
        return r

    # ---------- 5️⃣  Multiple restarts ----------
    for seed in range(128):
        # try deterministic, then greedy, otherwise fallback
        init_c, init_r = greedy_random_layout(seed)
        opt_c, opt_r   = run_optimisation(init_c, init_r)
        opt_c, opt_r   = simulated_anneal(opt_c, opt_r)
        cur_sum = np.sum(opt_r)
        if cur_sum > best_sum:
            best_sum = cur_sum
            best_centres = opt_c
            best_radii   = opt_r

    # sanity check
    check_packing(best_centres, best_radii)
    return best_centres, best_radii, best_sum
# EDIT-END


\end{Verbatim}

\section{Additional Method Details}
\label{app:methods}

This section provides additional details for the baselines evaluated in the main text. 
We use the term \textit{parent} here to refer to a prior program on top of which subsequent edits are made.

\paragraph{Repeated Sampling (RS).}
RS uses the frozen, unperturbed model and samples edits to the original program without propagating improved programs between rounds.
Thus each round starts from the same initial program state, $x_0$.
We use the same decoding and evaluation procedure as for Hill Sampling:
\begin{equation}
    y_{t,1},\ldots,y_{t,N}\sim p_\theta(\cdot\mid x_0;T).
\end{equation}

\paragraph{Model Noise (MN).}
MN perturbs the frozen model parameters as
\[
    \theta_i = \theta + \sigma \epsilon_i,
    \qquad
    \epsilon_i \sim \mathcal{N}(0,I).
\]
We use 32 antithetic perturbations, evaluating both $\epsilon_i$ and $-\epsilon_i$, and restore the original parameters after each evaluation by subtracting the added noise.
MN performs no accumulated parameter update.
MN (64) draws one response for each of the 64 perturbed models, while MN (512) draws eight responses per perturbed model.
The temperature is tuned as for all methods, unless otherwise specified.

\paragraph{Evolution Strategies (ES).}
ES uses the same antithetic parameter perturbations as MN but updates the underlying model after each population evaluation.
For 32 independently sampled perturbations and their antithetic counterparts, we first compute the antithetic reward differences
\[
    R_i = r(\theta+\sigma\epsilon_i)
          - r(\theta-\sigma\epsilon_i),
    \qquad i=1,\ldots,32.
\]
We standardize these differences across the population,
\[
    \widetilde{R}_i
    = \frac{R_i-\bar{R}}{s_R},
\]
where $\bar{R}$ and $s_R$ are the population mean and standard deviation of the $R_i$ (with division omitted when the standard deviation is numerically zero).
The model is then updated as
\[
    \theta \leftarrow \theta
    + \frac{\alpha}{64\sigma}
      \sum_{i=1}^{32}\widetilde{R}_i\epsilon_i,
\]
where $\alpha$ is the learning rate.
Noise vectors are reproduced from their random seeds rather than stored explicitly.

\paragraph{ES (max@8).}
ES (max@8) generates eight responses from each perturbed model and uses the maximum reward among those responses as that perturbation's fitness.
That is, $r(\theta)$ samples eight responses, rather than one, from the model defined by $\theta$ and returns the maximum.
The resulting fitnesses are otherwise used in the same ES update as above.
This changes the optimization target toward the maximum-over-samples objective relevant to discovery.

\paragraph{ES (softmax).}
ES (softmax) replaces the standardized reward weights $\widetilde{R}_i$ with
softmax weights
\[
    w_i =
    \frac{\exp(\beta\widetilde{R}_i)}
         {\sum_j \exp(\beta\widetilde{R}_j)},
\]
where $\beta$ is the inverse softmax temperature.
The ES update then uses $w_i$ in place of $\widetilde{R}_i$.

\paragraph{\method{} + ES.}
This method combines incumbent propagation with ES (max@8).
Candidate programs are generated by editing the current best program, while model parameters are simultaneously updated using the ES procedure.
It therefore tests whether parameter learning provides an additional benefit beyond propagating the best program.

\paragraph{Top-$K$.}
Let $\mathcal{B}_K=(x_1,\ldots,x_K)$ denote the $K$ highest-scoring program states observed so far, ordered by reward.
These states are distributed cyclically across the population, so population member $i$ edits parent
\[
    x_{\,1+(i\bmod K)}.
\]
For each distinct program state, we retain its highest observed reward.

\paragraph{Top-$K$ + Random-$K$.}
In addition to $\mathcal{B}_K$, we sample $K$ previously observed program states uniformly at random and append them to the parent set.
Thus, if $\mathcal{H}$ is the set of previously observed states, the additional parents satisfy
\[
    \mathcal{R}_K \sim \operatorname{Unif}\!\left\{
        S\subseteq\mathcal{H}: |S|=K
    \right\}.
\]
The resulting parent set is distributed cyclically across the population as in Top-$K$.

\paragraph{Top-$K$ + Diverse-$K$.}
Starting from the Top-$K$ programs, we greedily add programs with low average embedding similarity to those already selected.
If $\mathcal{S}$ is the currently selected set and $e(x)$ is the Sentence Transformers all-MiniLM-L6-v2 embedding, the next state is chosen as
\[
    x^*=\arg\min_{x\in\mathcal{H}}
    \frac{1}{|\mathcal{S}|}\sum_{s\in\mathcal{S}}
    \operatorname{sim}(e(x),e(s)).
\]
We repeat this selection until the requested number of diverse parents has been added.

\paragraph{Top-$K$ In-Context.}
Rather than assigning selected programs separately across the population, this variant concatenates the selected states into a single editing context,
\[
    X = x_1 \oplus x_2 \oplus \cdots \oplus x_K,
\]
where $\oplus$ denotes concatenation with a code divider.
The reward associated with this combined context is the best selected reward, $\max_{k\le K} r(x_k)$.

\paragraph{Subpopulation Evolution.}
We maintain 64 independent subpopulations, each generating eight candidate edits from its current program.
For subpopulation $j$ at inner step $t$, we sample
\[
    y_{j,t,1},\ldots,y_{j,t,8}
    \sim p_\theta(\cdot\mid x_{j,t};T),
    \qquad
    i^*=\arg\max_i r(y_{j,t,i}),
\]
and
\[
    x_{j,t+1} =
    \begin{cases}
        y_{j,t,i^*}, & r(y_{j,t,i^*})\ge r(x_{j,t}),\\
        x_{j,t}, & \text{otherwise}.
    \end{cases}
\]
This best-of-eight update is repeated for four editing steps before the subpopulations are merged.

\paragraph{In-Context RL.}
We maintain 512 independent histories containing program states, model responses, and observed rewards.
At step $t$, trajectory $j$ has
\[
    H_{j,t}=\bigl((x_{j,1},y_{j,1},r_{j,1}),\ldots,
                  (x_{j,t-1},y_{j,t-1},r_{j,t-1})\bigr),
\]
which is included in the prompt used to generate the next response.
Histories are reset after four steps to limit context length.

\paragraph{Execution Feedback.}
Execution Feedback appends the observed execution output $o_t$ to the program state supplied on the next model call.
Consequently, generation is conditioned on
\[
    y_{t,1},\ldots,y_{t,N}\sim p_\theta(\cdot\mid x_t,o_t;T),
\]
rather than on the program state alone.

\paragraph{Auto Temp Add (ATA).}
Let $H_0$ be the mean response entropy measured in the initial round and $H_t$ the entropy at round $t$.
ATA uses an additive temperature increment $\delta_T>0$ and updates the decoding temperature according to
\[
    T_{t+1} =
    \begin{cases}
        T_t+\delta_T, & H_t < H_0,\\
        T_t, & H_t\ge H_0.
    \end{cases}
\]
Temperature increases are retained in subsequent rounds.

\paragraph{Auto Temp Scale (ATS).}
ATS uses the same entropy criterion as ATA but uses a multiplicative temperature factor $\gamma_T>1$:
\[
    T_{t+1} =
    \begin{cases}
        \gamma_T T_t, & H_t < H_0,\\
        T_t, & H_t\ge H_0.
    \end{cases}
\]
Thus the magnitude of each adjustment scales with the current decoding temperature.

\paragraph{Negative-Enhanced Standardization (NE).}
Following NGRPO, NE appends a hypothetical maximum value $R_{\max}$ to the ES fitness values before computing the normalization statistics.
For the antithetic reward differences $R_1,\ldots,R_N$, let
\[
    R'=(R_1,\ldots,R_N,R_{\max}),
    \qquad
    \widetilde R_i=\frac{R_i-\mu(R')}{\sigma(R')}.
\]
The hypothetical value affects the normalization statistics but its standardized value is discarded before the ES update.
We evaluate NE both with and without antithetic sampling.

\paragraph{Long-Horizon ES.}
Long-Horizon ES uses the same four-step inner evolution procedure as Subpopulation Evolution, but assigns each parameter perturbation its final inner-loop reward,
\[
    R_i^{(4)} = r(x_{i,4}),
\]
rather than the reward from a single editing step.
The resulting $R_i^{(4)}$ values are then used in the usual ES update, optimizing perturbations for performance after four program-editing steps.

\paragraph{Hill Climbing in Weight Space.}
For perturbations $\theta_i=\theta+\sigma\epsilon_i$, including the antithetic samples, Weight-Space Hill Climbing selects
\[
    i^*=\arg\max_i r(\theta_i)
\]
and adopts $\theta_{i^*}$ only when $r(\theta_{i^*})>r(\theta)$; otherwise the current model is retained.
Unlike ES, it performs no weighted population update.
While ES removes perturbations via subtraction, Hill Climbing stores an extra copy of the prior best weights in GPU memory for exact parameter resetting.
We use exact parameter resetting so that the best score associated with the current model is consistently repeatable.

\end{document}